\documentclass[11pt]{article}

\usepackage[letterpaper,margin=1.25in]{geometry}
\usepackage{newtxtext,newtxmath}   %

\usepackage{amsmath,amssymb}
\usepackage{booktabs}
\usepackage{multirow}

\usepackage{graphicx}
\usepackage{xcolor}                %
\usepackage[font=small,labelfont=bf]{caption}
\usepackage{subcaption}
\usepackage{wrapfig}
\usepackage{float}
\usepackage{placeins}

\author{%
  Pegah Khayatan\textsuperscript{$\star$1} \quad
  Sara Meziane\textsuperscript{$\star$1} \quad
  Jayneel Parekh\textsuperscript{$\star$1} \quad
  Matthieu Cord\textsuperscript{1,2} \\[0.6em]
  \textsuperscript{1}ISIR, Sorbonne Université, Paris, France \quad
  \textsuperscript{2}Valeo.ai, Paris, France \\[0.3em]
  \texttt{\{pegah.khayatan, sara.meziane, jayneel.parekh\}@sorbonne-universite.fr}
}

\usepackage[numbers,sort&compress]{natbib}

\usepackage{xr}
\usepackage[accsupp]{axessibility}

\usepackage[colorlinks,citecolor=blue,linkcolor=blue,urlcolor=magenta]{hyperref}
\usepackage{cleveref} 
\usepackage{orcidlink}
\usepackage{underscore}
\usepackage{titling}
\usepackage{xspace}

\title{\textit{DiMaS}: \textit{Di}stribution \textit{Ma}tching for \\
       \textit{S}teering Vision-Language-Action Models}

\author{%
  Pegah Khayatan\textsuperscript{$\star$1} \quad
  Sara Meziane\textsuperscript{$\star$1} \quad
  Jayneel Parekh\textsuperscript{$\star$1} \quad
  Matthieu Cord\textsuperscript{1,2} \\[0.6em]
  \textsuperscript{1}ISIR, Sorbonne Université, Paris, France \quad
  \textsuperscript{2}Valeo.ai, Paris, France \\[0.3em]
  \texttt{\{pegah.khayatan, sara.meziane\}@isir.upmc.fr}
}
\date{}

\begin{document}
\maketitle
\renewcommand{\thefootnote}{\fnsymbol{footnote}}
\footnotetext[1]{Equal contribution.}
\renewcommand{\thefootnote}{\arabic{footnote}}
\raggedbottom

\noindent\textbf{Abstract:} Flow-matching-based vision-language-action (VLA) models have emerged as powerful policies for robotic manipulation, yet a critical capability remains underexplored: fine-grained behavioral control, the ability to govern \emph{how} a robot performs a task by intervening on its internal representations. Representation steering is a well-established interpretability tool for language and vision-language models, where behavioral features are typically encoded as linear directions, but we show that these classic methods fall short in VLAs. We propose \textbf{DiMaS}, a \textbf{Di}stribution-\textbf{Ma}tching \textbf{S}teering strategy tailored to flow-matching VLAs, which transports between representation distributions rather than shifting along a fixed direction, and show that it effectively controls behavior across two state-of-the-art VLAs. We further examine the generalizability of this strategy as the tasks it is learned from and evaluated on grow increasingly dissimilar, characterizing where behavioral control transfers and where it weakens. Finally, through an analysis of the representation structure of the action expert, we explain why classical linear steering falls short in the visuomotor setting: behavioral features are linearly decodable but not linearly steerable, which motivates the distribution-matching design of DiMaS. Our code is publicly available.\footnote[1]{Github page: \url{https://github.com/pegah-kh/dimas}}$^,$ \footnote[2]{Blog/Project page: \url{https://pegah-kh.github.io/dimas/}}

\vspace{0.5em}
\noindent\textbf{Keywords:} Vision-Language-Action Models; Behavioral Control; Representation Steering; Robot Manipulation

\newcommand{\bmX}{\mathcal{X}}
\newcommand{\bmY}{\mathcal{Y}}
\newcommand{\bmS}{\mathcal{S}}
\newcommand{\xI}{x^{\mathcal{I}}}
\newcommand{\bmF}{\mathcal F}
\newcommand{\bmG}{\mathcal G}
\newcommand{\bmGF}{{\mathcal G}_{\bmF}}
\newcommand{\bmI}{\mathcal I}
\newcommand{\bmL}{\mathcal L}
\newcommand{\bmD}{\mathcal D}

\newcommand{\argmin}{\mathop{\rm argmin}}
\newcommand{\argmax}{\mathop{\rm argmax}}

\definecolor{gold2}{rgb}{0.65,0.55,0.01}

\newcommand{\bfT}{\mathbf {T}}
\newcommand{\bfX}{\mathbf {X}}
\newcommand{\bfZ}{\mathbf {Z}}
\newcommand{\bfU}{\mathbf {U}}
\newcommand{\bfV}{\mathbf {V}}
\newcommand{\bfR}{\mathbf {R}}

\newcommand{\reals}{\mathbb{R}}
\newcommand{\Va}{{\mathbf{X}}}
\newcommand{\Ha}{{\mathbf{H}}}
\newcommand{\Haint}{{\mathbf{H}}}
\newcommand{\Hak}[1]{{\mathbf{h}_{#1}}}
\newcommand{\Wa}{{\mathbf{W}}}
\newcommand{\Wak}[1]{{\mathbf{w}_{#1}}}

\section{Introduction}

The recent rise of foundation models, spanning Large Language Models
(LLMs)~\cite{brown2020languagegpt3, touvron2023llama} and Vision-Language Models (VLMs)~\cite{radford2021learning, alayrac2022flamingo},
has driven rapid progress in robot policies, now commonly framed as Vision-Language-Action (VLA) models~\cite{zitkovich2023rt, kim2024openvla, shukor2025smolvla, intelligence2025pi05, fang2026molmoact2}. These policies aim to transfer the rich visual and linguistic priors encoded in LLMs and VLMs to robotic control.

Concretely, a VLA takes images from visual sensors together with a natural-language
task instruction as multimodal inputs, and predicts low-level robot actions; these
inputs are typically processed by a pretrained or finetuned VLM backbone. Early VLAs
generated actions autoregressively, treating them as discrete
tokens~\cite{zitkovich2023rt, kim2024openvla}. More recent state-of-the-art systems
instead pair the VLM with a flow-matching~\cite{lipman2022flow} action expert that
predicts continuous actions conditioned on the VLM
representations~\cite{intelligence2025pi05, shukor2025smolvla}.

Despite this progress, \textit{steering} VLAs remains underexplored: that is,
developing \textit{post-hoc} methods that control specific attributes of the predicted
actions by intervening on internal representations. For any deployed generative model, such control carries substantial application value, since it
gives a user a direct mechanism to shape the output. Representation steering is now a well-established and effective tool for LLMs and
VLMs~\cite{turner2023steering, zou2023representation, khayatan2025analyzing, parekh2026learning}, yet it
is unclear how these recipes carry over to VLAs, where a
flow-matching action expert produces continuous trajectories rather than discrete tokens.
Our goal in this paper is to develop steering methods that intervene on VLA representations to control targeted features of the predicted trajectory, and to verify that this control generalizes to settings beyond those used to calibrate the intervention. Further, our analysis of the action expert's representation structure yields broad insights into why classical steering strategies fall short in VLAs.

Through this work, we make key contributions in the following directions:
\begin{itemize}
    \item We propose a simple and novel \textbf{Di}stribution \textbf{Ma}tching based \textbf{S}teering (DiMaS) mechanism, a steering strategy specifically adapted to recent flow-matching VLAs. It intervenes on latent representations and applies across models of different scales, which we validate on SmolVLA~\cite{shukor2025smolvla} and
    $\pi_{0.5}$~\cite{intelligence2025pi05} across multiple steering tasks.

    \item 
    We examine the generalizability of steering as the tasks it is learned from and evaluated on grow increasingly dissimilar, characterizing where behavioral control transfers and where it weakens, and yielding insight into the robustness of representation-level control in VLAs.

    \item Through an investigation of representation structure inside a flow-matching VLA, with a particular focus on the action expert, we provide insights into \textit{why} classical linear steering strategies that succeed for LLMs and VLMs fall short in the visuomotor setting, motivating the design of DiMaS.

\end{itemize}

\section{Related Works}

\noindent\textbf{Representation steering in generative models.}
Representation steering refers to performing inference-time interventions on the internal representations of generative models to control specific attributes or behaviors of their output. With roots in latent-space editing of VAEs and GANs~\cite{vae_2013, shen2020interfacegan},
the idea has gained significant traction with the rise of large language models (LLMs) and multimodal large language models (MLLMs), where it has proven effective for addressing alignment problems \cite{turner2023steering, panickssery2023steering, arditi2024refusal, qiu2025hallucination} and controlling generation style in multimodal models \cite{khayatan2025analyzing, parekh2026learning, pach2026sparse}. \newline
Most such approaches are grounded in the \textit{linear representation hypothesis} (LRH), which posits that semantic features are encoded as linear directions in activation space~\cite{park2023linear}, and that representations can be expressed as sparse linear combinations of such directions~\cite{elhage2022toy}, enabling simple additive interventions at inference time. While recent work has questioned the LRH~\cite{engels2025not, fel2025rabbit} and proposed richer models of concept representation~\cite{bhalla2026sparse}, linear steering remains the dominant paradigm for controlling LLMs and VLMs. We find that a fixed linear shift is inadequate for controlling behavioral features in VLAs, motivating a strategy based on distribution matching. In spirit, our approach is most related to the optimal transport-based steering of Rodriguez~et~al.~\cite{rodriguez2025controlling}, which they develop for LLMs and diffusion models; our focus differs, targeting behavioral control in the visuomotor representations of flow-matching VLAs. \newline

\noindent\textbf{Interpretability and steering for VLAs.} \citet{haon2025mechanistic} study mechanistic interpretability in autoregressive VLAs that generate discrete action tokens. They identify clusters of FFN value vectors that can be steered at inference time by overriding the corresponding neuron activations, and give early evidence that such interventions can modulate motion speed and trajectory height. Their framework applies only to autoregressive VLAs, whereas we target the current state-of-the-art flow-matching VLAs.

Closest to our work, \citet{buurmeijer2026observing} formalize \textit{feature-observability} and \textit{feature-controllability} in VLAs. They learn a regressor to predict behavioral features from VLM hidden states and use its coefficients as a steering vector to shift the VLM
representations at inference time. A key limitation is that they must pause the intervention to
preserve task success.
Our method preserves task success without pausing the intervention, and replaces the linear shift with a steering strategy based on distribution matching.

\section{Method}

We first introduce the notation and the general VLA architecture we consider in
\Cref{method:notation}, and then present our method, DiMaS, in \Cref{method:dimas}.

\subsection{Notation and VLA architecture}
\label{method:notation}
We assume a generic VLA architecture that covers the paradigm of current SOTA VLAs based on joint VLM and flow-matching networks for action prediction. At any timestep $t$, the VLA $f$ receives an observation $x_t = (I_t, T, x_t^s)$ consisting of an image, text instruction, and proprioceptive state and predicts low-level actions for each robot joint, denoted as $y_t$. The VLA $f = (f_V, f_A)$ consists of two main components, $f_V$, which is the VLM backbone and $f_A$, the action expert network. The VLM representations are used to condition the action expert $f_A$ that generates the actions. It is trained with a flow-matching objective \cite{lipman2022flow} to iteratively refine noisy action tokens $a_{0,t} \sim \mathcal{N}(0, \mathbf{I})$ across $M$ denoising steps. 
\begin{equation}
    a_{m+1, t} = a_{m,t} + \frac{1}{M}\, f_A\!\left(a_{m,t},\, m,\, f_V(x_t)\right), \, \,  a_{0,t} \sim \mathcal{N}(0, \mathbf{I}),\ m \in \{0, \ldots, M-1\}
\end{equation}
The final predicted action chunk $y_t$ is obtained directly from $a_{M,t}$ through an environment-specific transformation.

Throughout this paper, we analyze or intervene on residual stream representations that can be sourced from different parts of $f$. In the LLM backbone of the VLM, the residual stream representation at layer $l$ and token position $p$ is denoted as $h^p_l(x_t) \in \mathbb{R}^{d_V}$. For the action expert $f_A$, analogously, they are denoted as $h^{p,m}_l(x_t) \in \mathbb{R}^{d_A}$, where additionally, $m$ is the denoising step. %

\subsection{\textit{DiMaS}: Distribution matching for steering VLA representations}\label{method}

\begin{figure}[t]
    \centering
    \includegraphics[width=\linewidth]{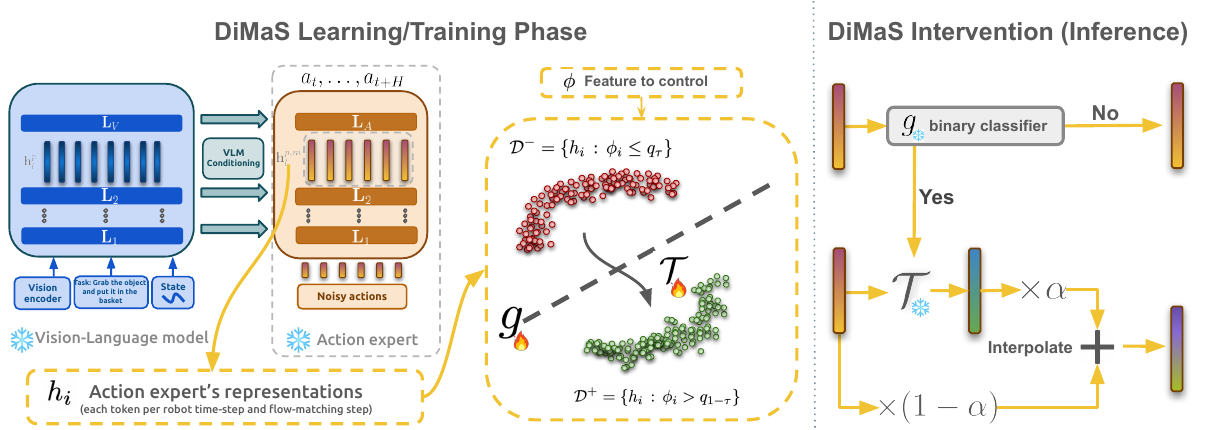}
    \caption{\textbf{DiMaS Method Overview}: A VLM backbone encodes multimodal robot inputs and conditions a flow-matching based action expert to predict the final action chunk. \textbf{Training:} We extract residual-stream representations $h^{p,m}_{l}$ from the action expert at a specific layer $l^*$, and flow-matching step $m$. After obtaining a continuous behavioral feature (e.g.\ speed); the representations are grouped by feature value to create source and target distributions $\mathcal{D}^{-}$, $\mathcal{D}^{+}$ (feature absent, present respectively). The steering intervention is learnt as a transport map between the two distributions. \textbf{Inference:} During test-time, steering is gated via a binary classifier $g$. To achieve feature control and high success rate we interpolate between the source/target distributions through the transport map $\mathcal{T}$ and interpolation knob $\alpha \in [0, 1]$. %
    }
    \label{fig:architecture}
\end{figure}

We introduce a novel steering method grounded in optimal transport that learns a
mapping between distributions of internal representations. A complete overview of the method is illustrated in \Cref{fig:architecture}. \newline
\noindent\textbf{Representations and behavioral feature.}
The first step of our method sources representations from action expert $f_A$, and pairs each with a continuous scalar feature/concept measuring a target behavioral property we wish to control.\newline
For a given network layer $\ell$ and flow-matching denoising step $m$, we extract representations $h^{p,m}_{\ell}(x_t) \in \mathbb{R}^d$ for each $x_t$ across all rollouts. Each position $p$ yields a representation. For simplicity of notation, we will denote the $i$'th extracted representation as $h_i$. To extract the corresponding scalar feature in our experiments, we use the predicted action to extract the feature of interest (although not necessary for our method). Concretely, each predicted action is a vector
$a_i = (\Delta x, \Delta y, \Delta z, \dots, \texttt{gripper})$, encoding the
end-effector displacement together with the gripper state. From these actions we derive
a scalar behavioral feature; for instance, when the targeted property is the end-effector \emph{speed}, the feature is the magnitude of the translational
displacement $\phi(a_i) = \sqrt{\Delta x^2 + \Delta y^2 + \Delta z^2}$. Each representation $h_i$ is then associated with the feature value of its corresponding action $\phi_i=\phi(a_i)$, giving a continuous score for the behavioral property we wish to control. \newline
\noindent\textbf{Source and target distributions.}
To partition the representations into a source distribution $\mathcal{D}^{-}$ and a
target distribution $\mathcal{D}^{+}$, we threshold on the behavioral feature using its empirical quantiles: 

\begin{equation}
    \mathcal{D}^{-} = \{h_i\,:\, \phi_i \leq q_{\tau}\}, \, \, \mathcal{D}^{+} = \{h_i\,:\, \phi_i > q_{1-\tau}\}
\end{equation}
Representations whose feature value falls below the lower quantile $q_{\tau}$ form the source set $\mathcal{D}^{-}$, while those above the upper quantile $q_{1-\tau}$ form the target set $\mathcal{D}^{+}$. Restricting each distribution to a tail, rather than splitting at the median, yields cleaner, well-separated populations of the feature-absent and feature-present behavior, which sharpens the learned mapping.

\noindent\textbf{Steering as a transport map.}
At its core, steering a behavioral feature amounts to learning a map $\mathcal{T}$ that transports a representation from the source distribution $\mathcal{D}^{-}$ to the target distribution $\mathcal{D}^{+}$, thereby inducing the feature in representations that lack it. Different steering strategies correspond to different choices of $\mathcal{T}$. The simplest and most widely used is \emph{linear} steering, where $\mathcal{T}$ is an additive shift in a fixed direction. This fixed direction can be extracted in multiple ways. For instance, one popular form is the \emph{mean-difference} steering, where this direction is equal to the difference of the distribution means,
$\mathcal{T}(h) = h + (\mu^{+} - \mu^{-})$, with $\mu^{\pm}= \mathbb{E}_{h \sim \mathcal{D}^\pm}[h]$. A more flexible variant of linear steering is based on \emph{regression}, where $\mathcal{T}$ displaces the
representation along the direction of a regressor trained to predict the feature value,
so that the shifted representation attains a desired target value (refer to Appendix~\ref{app:baselines} for more details). Both are linear by construction, which, as we empirically show in \Cref{analysis:representation_structure}, limits their ability to control behavioral features in VLAs with a denoising action expert. DiMaS instead instantiates $\mathcal{T}$ as an \emph{optimal transport} map,
which respects the full geometry of $\mathcal{D}^{-}$ and $\mathcal{D}^{+}$ rather than a
single direction.

Concretely, we learn the transport map $\mathcal{T}^{(m)}_{\ell}(\cdot)$ from $\mathcal{D}^{-}$ to
$\mathcal{D}^{+}$ (see Figure~\ref{fig:architecture}, left) by minimizing the Kantorovich optimal transport objective,
\begin{equation}
\mathcal{W}_2^2(\mathcal{D}^{-}, \mathcal{D}^{+})
= \min_{\gamma \in \Pi(\mathcal{D}^{-}, \mathcal{D}^{+})}
\int_{\mathcal{Z} \times \mathcal{Z}} \|z^{-} - z^{+}\|^2 \,
\mathrm{d}\gamma(z^{-}, z^{+}),
\end{equation}

where $\mathcal{Z}$ denotes the support of the distributions and $\|\cdot\|$ is the standard Euclidean norm acting as the ground cost. In practice, $\mathcal{D}^{-}$ and $\mathcal{D}^{+}$ are only accessible through finite empirical samples, so this reduces to a discrete optimal transport problem, which we solve efficiently via low-rank Sinkhorn~\cite{scetbon2021lowrank}; full details are provided in \Cref{app:dimas_ot_details}. \newline
\noindent\textbf{Test-time intervention strategy.}
Given a representation $h = h^{p,m}_{\ell}(x_t)$ at test time, we first project $h$ onto
$\mathcal{D}^{-}$ via a nearest-neighbor projection
$P(h) = \arg\min_{z \in \mathcal{D}^{-}} \|z - h\|$, and then apply the learned transport
map to obtain the steered representation $\mathcal{T}^{(m)}_{\ell} \circ P(h)$. To avoid
unnecessary interventions, steering is gated by a \textit{feature filter}: a binary
classifier $g_{\ell}^{(m)} : \mathbb{R}^{d} \to \{0, 1\}$, implemented as a linear probe
trained to separate $\mathcal{D}^{-}$ from $\mathcal{D}^{+}$. The gate becomes active, i.e., 
$g_{\ell}^{(m)}(h) = 1$ when the feature is \emph{absent}, i.e.\ $h$ is predicted to lie
in $\mathcal{D}^{-}$, and $g_{\ell}^{(m)}(h) = 0$ when it is already present, so that
representations already exhibiting the targeted feature are left unchanged. \newline
\noindent\textbf{Interpolation as the secret sauce.} A crucial goal behind our steering methodology is that a user should be able to control the feature of interest while simultaneously succeeding at the task. In other words, task success should not be achieved by needing to stop the steering intervention. In our experiments, we observed that while completely transporting $h$ to $\mathcal{D}^+$ via $\mathcal{T}^{(m)}_{\ell} \circ P(h)$ controls the feature of interest, it introduces deviation from the original $h$ that can lead to task failures. Given that we perform evaluations in a closed-loop setup and tasks/models can often be fragile, this sometimes causes significant drop in success rate. To ensure a minimal drop in success rate while still controlling the feature, we interpolate between the distributions $\mathcal{D}^-$ and $\mathcal{D}^+$, via a parameter $\alpha \in [0, 1]$. This intervention is summarized below and illustrated on the right of \Cref{fig:architecture}:

\begin{equation}
    h \;\leftarrow\;
    \begin{cases}
        (1 - \alpha)\, h \;+\; \alpha \, \big(\mathcal{T}^{(m)}_{\ell} \circ P(h)\big),
            & \text{if } g_{\ell}^{(m)}(h) = 1, \\[4pt]
        h, & \text{otherwise.}
    \end{cases}
\end{equation}
Here $\alpha = 0$ recovers the unsteered representation and $\alpha = 1$ applies the full transport, with intermediate values yielding a continuous modulation of the target behavior. From a user perspective, $\alpha$ provides a normalized, interpretable interpolation knob to control steering strength while maintaining success rate as much as possible. Interestingly, for OT based LLMs/VLM steering \cite{rodriguez2025controlling}, $\alpha$ plays a similar role of a normalized knob controlling steering strength. For DiMaS, it serves an even more important purpose by also affecting the success rate.%

\label{method:dimas}

\section{Experiments}

We evaluate our method on two flow-matching-based VLA models of different scales,
SmolVLA~\cite{shukor2025smolvla} and $\pi_{0.5}$~\cite{intelligence2025pi05} and LIBERO benchmark \cite{libero} for controlled, axis-isolated evaluation. This simulator consists of four suites, namely LIBERO-Object, LIBERO-Spatial, LIBERO-Goal,
and LIBERO-10, where the first three comprise shorter-horizon tasks and LIBERO-10 comprises
longer-horizon ones. Each suite contains 10 tasks, and each task provides 50 initializations that vary initial placement of the objects. More details are provided in Appendix~\ref{app:models_and_datasets}. \newline
\noindent\textbf{Target feature.}
We consider two target behavioural features: \textit{motion speed}, measured as the
mean norm of end-effector velocity, and \textit{end-effector vertical displacement}, measured as the
accumulated absolute displacement along the vertical axis. \newline
\noindent\textbf{Baselines.}
We compare against three categories of baselines.
\textit{(i) Mean-difference steering} applies a constant additive shift equal to the
difference of the source and target means, $\mathcal{T}(h) = h + (\mu^{+} - \mu^{-})$, where $h$ represents a hidden representation of the VLM component. \textit{(ii) Regression-based steering}, based on \cite{buurmeijer2026observing}, fits a linear regressor to predict $\phi$ from $h$ and shifts the representation along
the regressor's direction so that its predicted feature attains a target quantile $q^\star$. We evaluate this baseline both in the VLM and the flow-matching (FM) action expert. Its closed form is given in Appendix~\ref{app:baselines}. Both representation-level baselines are linear steering methods by construction, with the shift
gated by a linear classifier and applied only to representations that do not yet exhibit the
target behavior.
\textit{(iii) Prompt injection} modulates behavior purely through instruction rephrasing
(e.g.\ ``go faster,'' ``slower''), with no activation-level intervention. \newline
\noindent\textbf{Metrics.}
We report three metrics: \textit{Success Rate} (SR), which measures task completion; \textit{feature value}, the value of the targeted behavioral feature (speed or $z$-displacement); and the \textit{statistical significance} of the feature shift relative to the unsteered baseline, assessed via paired $t$-tests across episodes.

\subsection{Can representation steering control behavior?}
\begin{figure}
    \centering

    \vspace{0.3em}
    \hspace{-5em}
    \begin{minipage}[c]{0.1\linewidth}
        \centering
        \includegraphics[width=\linewidth]{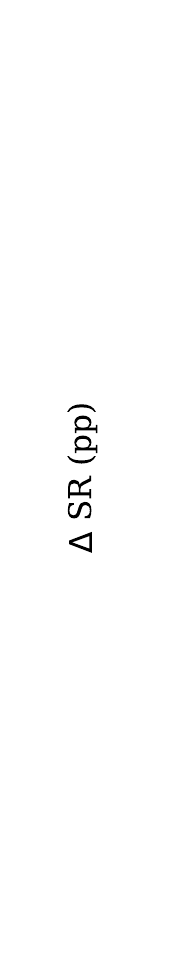}
    \end{minipage}%
    \hspace{0.1em}
    \begin{subfigure}[c]{0.4\linewidth}
        \centering
        \includegraphics[width=\linewidth]{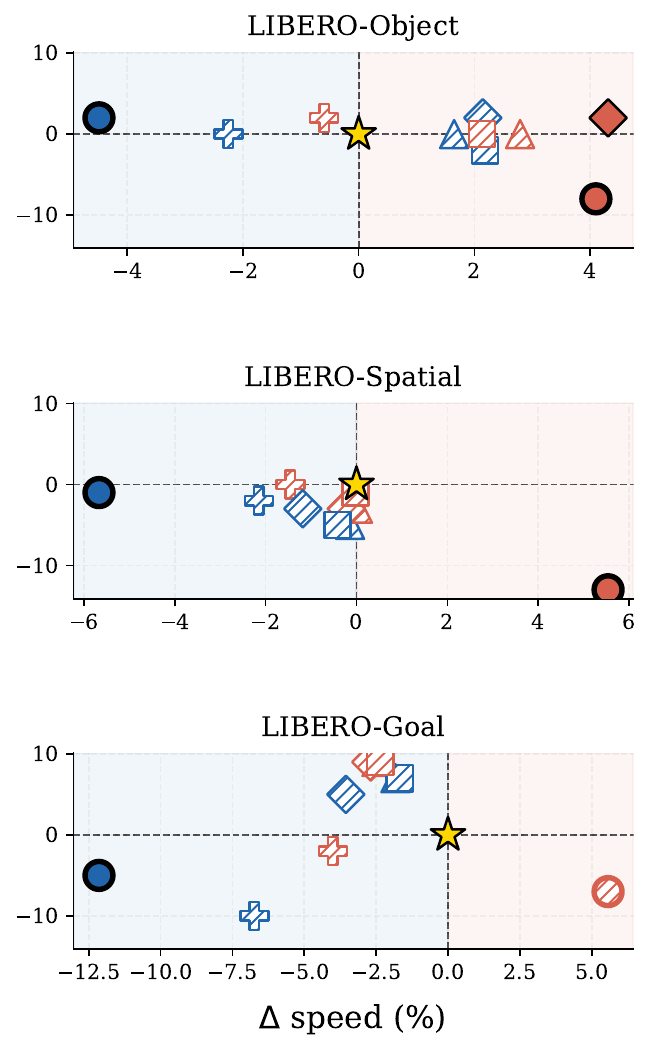}
        \subcaption{$\pi_{0.5}$}
    \end{subfigure}%
    \hspace{1em}
    \begin{subfigure}[c]{0.4\linewidth}
        \centering
        \includegraphics[width=\linewidth]{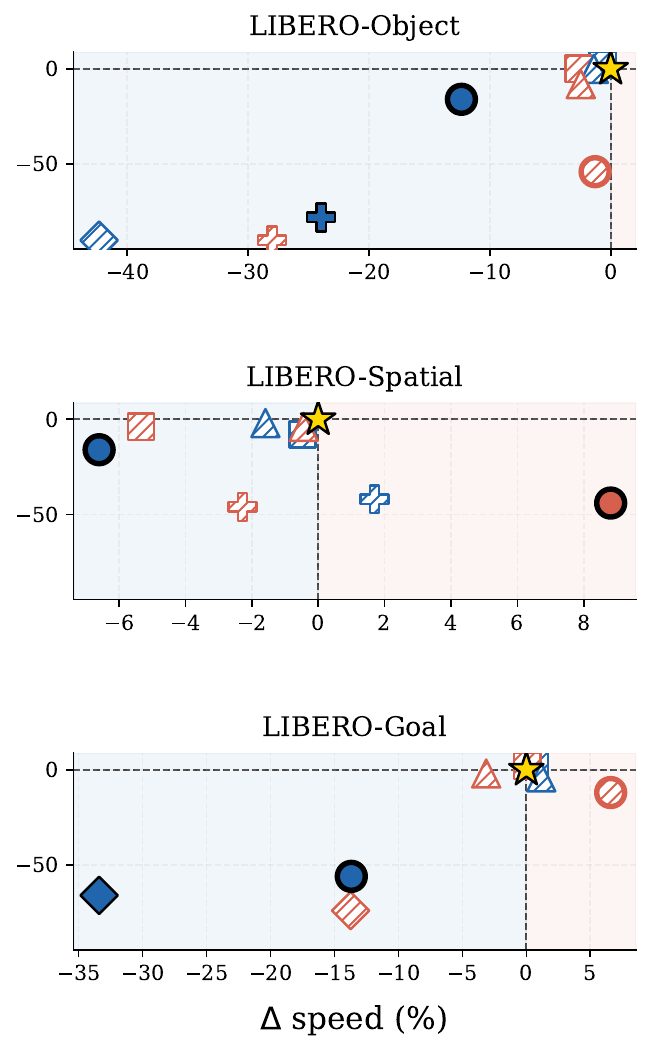}
        \subcaption{SmolVLA}
    \end{subfigure}

    \vspace{1em}
    \hspace*{0.0\linewidth}\hspace{0.1em}%
    \begin{minipage}{\dimexpr0.8\linewidth+2em\relax}
        \centering
        \includegraphics[width=0.5\linewidth]{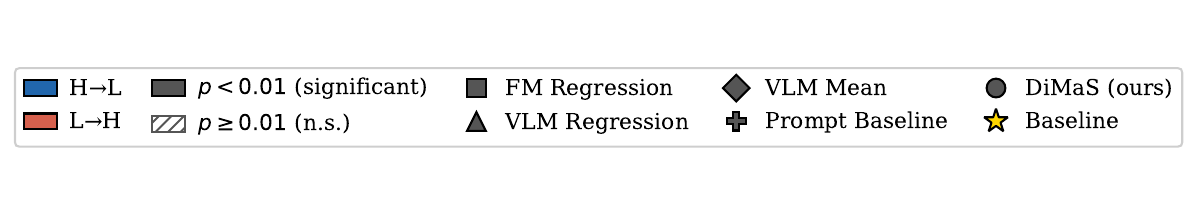}
    \end{minipage}

    \caption{\textbf{Speed steering.} Change in mean end-effector speed ($\Delta$speed) versus change in success rate ($\Delta$SR), relative to the unsteered baseline, across the three LIBERO suites, for (a) $\pi_{0.5}$ and (b) SmolVLA. Left of the vertical dashed line corresponds to high-to-low (H→L) steering, right to low-to-high (L→H). Marker shape denotes method; solid fill indicates a statistically significant change relative to baseline (t-test, p<0.01).}
    \label{fig:tradeoff_speed}
\end{figure}

\begin{figure}
    \centering

    \vspace{0.3em}
    \hspace{-5em}
    \begin{minipage}[c]{0.1\linewidth}
        \centering
        \includegraphics[width=\linewidth]{paper_figs/pi05_box_plots/baselines_figs/fig2_pi05_label.pdf}
    \end{minipage}%
    \hspace{0.1em}
    \begin{subfigure}[c]{0.4\linewidth}
        \centering
        \includegraphics[width=\linewidth]{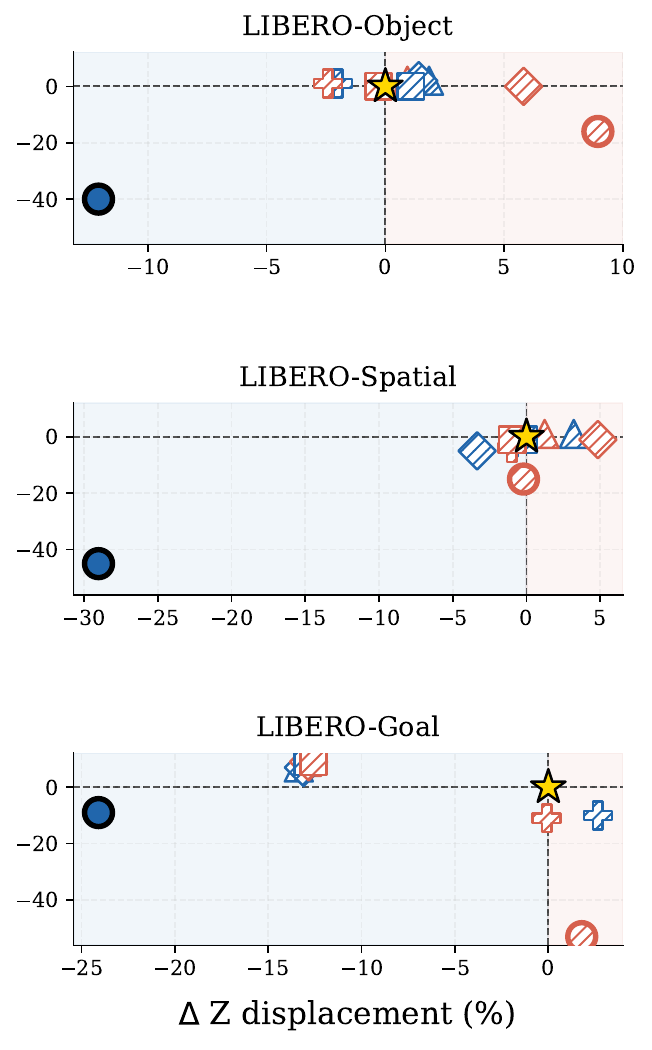}
        \subcaption{$\pi_{0.5}$}
    \end{subfigure}%
    \hspace{1em}
    \begin{subfigure}[c]{0.4\linewidth}
        \centering
        \includegraphics[width=\linewidth]{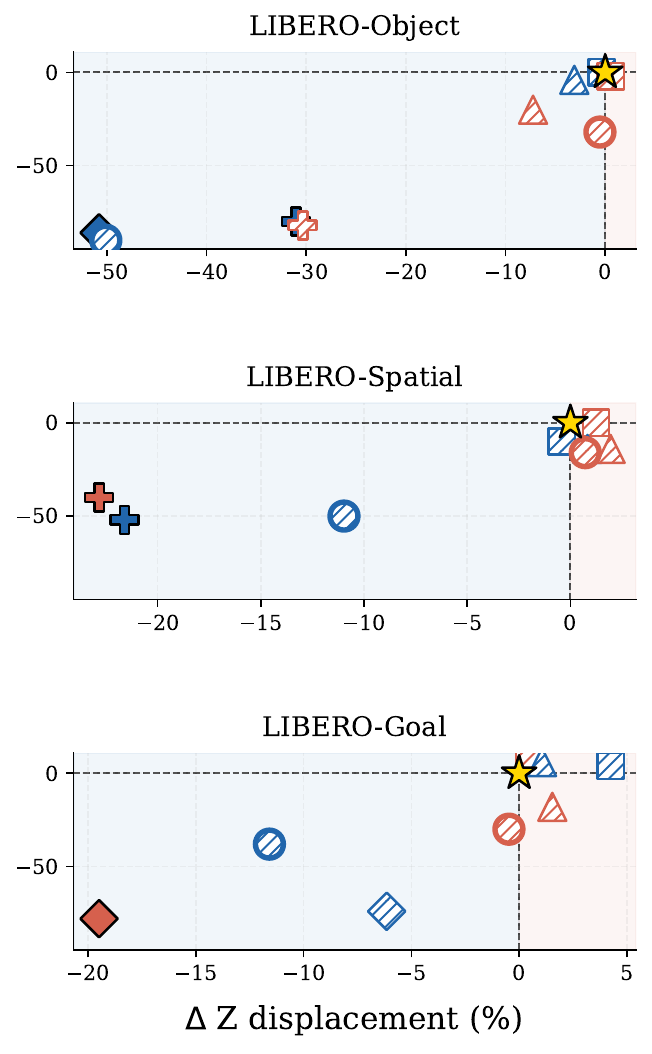}
        \subcaption{SmolVLA}
    \end{subfigure}

    \vspace{1em}
    \hspace*{0.0\linewidth}\hspace{0.1em}%
    \begin{minipage}{\dimexpr0.8\linewidth+2em\relax}
        \centering
        \includegraphics[width=0.5\linewidth]{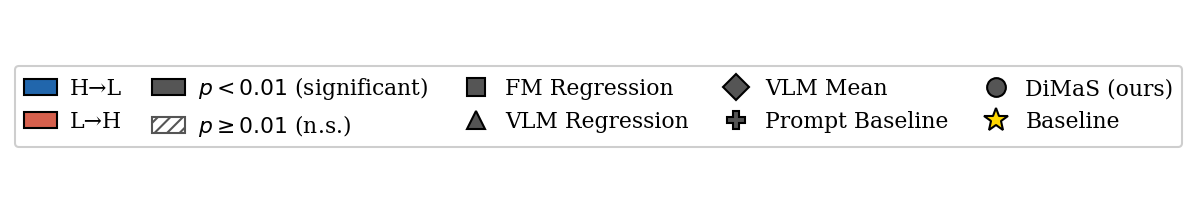}
    \end{minipage}

    \caption{\textbf{Z-displacement steering.} Change in accumulated vertical displacement ($\Delta$z-displacement) versus change in success rate ($\Delta$SR), relative to the unsteered baseline, across the three LIBERO suites. Conventions as in \Cref{fig:tradeoff_speed}.}
    \label{fig:tradeoff_height}
\end{figure}

We first ask whether the targeted feature can be modulated/controlled, and how DiMaS compares to the classic steering and instruction-level baselines introduced above. For each method we measure two quantities relative to the unsteered policy: whether it shifts the mean feature value (paired $t$-test) and how much its success rate changes. \Cref{fig:tradeoff_speed} and \Cref{fig:tradeoff_height} present these results for decreasing (blue, H$\to$L, high to low) and increasing (red, L$\to$H, low to high) end-effector speed and vertical displacement, evaluated across all baselines. Statistically significant shifts are shown in solid fill, while non-significant ones are striped. \newline
The linear steering and prompt baselines perform inconsistently. They leave the target feature unchanged, and where they do move it, they do so unreliably: the increasing and decreasing interventions often shift the feature in the same direction rather than opposite ones. DiMaS, by contrast, modulates the target feature in both directions: speed on both models, and vertical displacement on $\pi_{0.5}$. We largely preserve the success rate when modulating speed, but incur a drop when modulating vertical displacement. This is an expected result, since this feature is often tied to task completion. On LIBERO-Object, for example, decreasing vertical displacement keeps the end-effector too low to lift the object into the basket, so the placing step fails. Finally, for both features the decreasing intervention produces the more statistically significant shifts (\emph{e.g.} color-filled marks).

While $\pi_{0.5}$ achieves the strongest results overall, our steering approach proves effective across both models despite their substantial difference in scale, underscoring its generality. SmolVLA shows comparatively more sensitivity to action intervention, which we attribute in part to its action chunking strategy: SmolVLA predicts a new chunk at every timestep, producing more dynamic, less smoothed trajectories, whereas $\pi_{0.5}$ predicts 10 actions at a time, yielding smoother motion that is inherently more robust to intervention.

\subsection{Understanding steering generalization}
\label{gen_des}
We next investigate what makes a learned steering effective, and how well it transfers, using LIBERO's axis-isolated task structure. We vary two axes: (1) task diversity, the number and variety of tasks the steering vector is learned from, and (2) test distribution shift severity, i.e. how much the evaluation tasks differ from the training tasks. 
\begin{enumerate}
    \item \textbf{Setting 1: Held-out initial states of the same task.} For each task individually, we learn a steering from a subset of initial states and evaluate it on the remaining, held-out initial states of the same task.
    \item \textbf{Setting 2: Held-out initial states across tasks in a suite.} Considering all tasks jointly, we learn a single steering from a subset of initial states and evaluate its transfer to the remaining initial states. This differs from Setting~1 only in the diversity of the tasks used to learn the steering, not in the evaluation protocol:
    both evaluate on held-out initial states, so the comparison isolates the effect of learning from a single task versus from many.
    \item \textbf{Setting 3: Held-out tasks within a suite.} Within a given LIBERO suite
    (e.g.\ LIBERO-Object), we learn a steering from a subset of tasks and evaluate it on the held-out tasks of the same suite. This is the first setting that tests transfer to tasks unseen at training time.
    \item \textbf{Setting 4: Transfer across disjoint suites.} We learn a steering from one LIBERO suite (e.g.\ LIBERO-Object) and evaluate it on a different, disjoint suite
    (e.g.\ LIBERO-Goal), testing transfer across distinct task distributions.
\end{enumerate}

Settings~1 and~2 differ in the diversity of the tasks the steering is learned from, asking how much that diversity matters. Settings~3 evaluates transferability of steering within the same suite (\emph{e.g.} LIBERO-object), while Settings~4 further stress tests the transfer to other suites (\emph{e.g.} LIBERO-goal). Together, these settings provide fine-grained evidence about both what the steering needs in order to work and how robustly it generalizes. It is worth emphasizing that most steering and even VLA evaluation on LIBERO is typically done in Setting 2. Our settings 3 and 4 validate increasingly out-of-distribution capabilities of the steering method. 
We compare these steering settings for decreasing the end-effector speed and illustrate the results in \Cref{fig:levels_comparison_h_to_l_speed} (comparison for other tasks is provided in Appendix~\ref{app:generalization}). \newline
Across both models, all three suites, and all four settings, \textsc{DiMaS} shifts speed in the intended direction, with a significant decrease in nearly every case: an intermediate-layer intervention reliably slows the executed action. We now compare the four settings, which progressively increase the gap between the tasks the steering is learned from and those it is evaluated on. \newline
\noindent\textbf{Aggregating tasks helps (Setting~1 vs.\ Setting~2).}
Settings~1 and~2 share the same evaluation protocol and differ only in the diversity of the
tasks used to learn the steering. Learning from many tasks jointly (Setting~2) yields a more
consistent effect than learning from a single task (Setting~1): the induced shift is more
stable and the effect is significant across suites on both models. We attribute this to the
breadth of the learning set: a single task samples a narrow region of representation space,
so the learned transport partly reflects features specific to that task, whereas aggregating
tasks lets it capture the structure associated with the target behavior across a broader
region of the representation space. \newline
\noindent\textbf{Steering persists on held-out and disjoint tasks (Settings~3--4).} Importantly, DiMaS control effectiveness does not collapse when the steering is applied to tasks unseen at training time (Setting~3) or transferred across disjoint suites (Setting~4). For $\pi_{0.5}$, decreasing speed remains significant across suites at the held-out-task setting (Setting~3), and the effect persists in the intended direction even under the hardest cross-suite transfer to Goal, where the mean shifts in the intended direction although the change does not reach significance at $p<0.01$ (Setting~4, $p=0.032$).
The learned transport therefore captures structure tied to the target behavior rather than to the specific tasks it was learned from, and remains usable beyond its training setting. \newline
\noindent\textbf{Effectiveness is weakest on the most diverse suite.} LIBERO-Goal is the hardest suite. Its tasks are behaviorally diverse, spanning pushing objects and operating a stove in addition to pick-and-place, whereas Object and Spatial share a pick-and-place behavior across all tasks. This diversity coincides with weaker steering: Goal is the only suite with cells that do not reach significance at $p<0.01$ (SmolVLA Setting~1, $p=0.09$; $\pi_{0.5}$ Setting~4, $p=0.032$), and its shifts are generally smaller and noisier than on Object or Spatial, which are significant at every setting on both models. Both Goal's task heterogeneity and the smaller per-task sample available at Setting~1 likely contribute to this effect.

\begin{figure}[h]
    \centering

    \begin{subfigure}{0.48\linewidth}
        \centering
        \includegraphics[width=\linewidth]{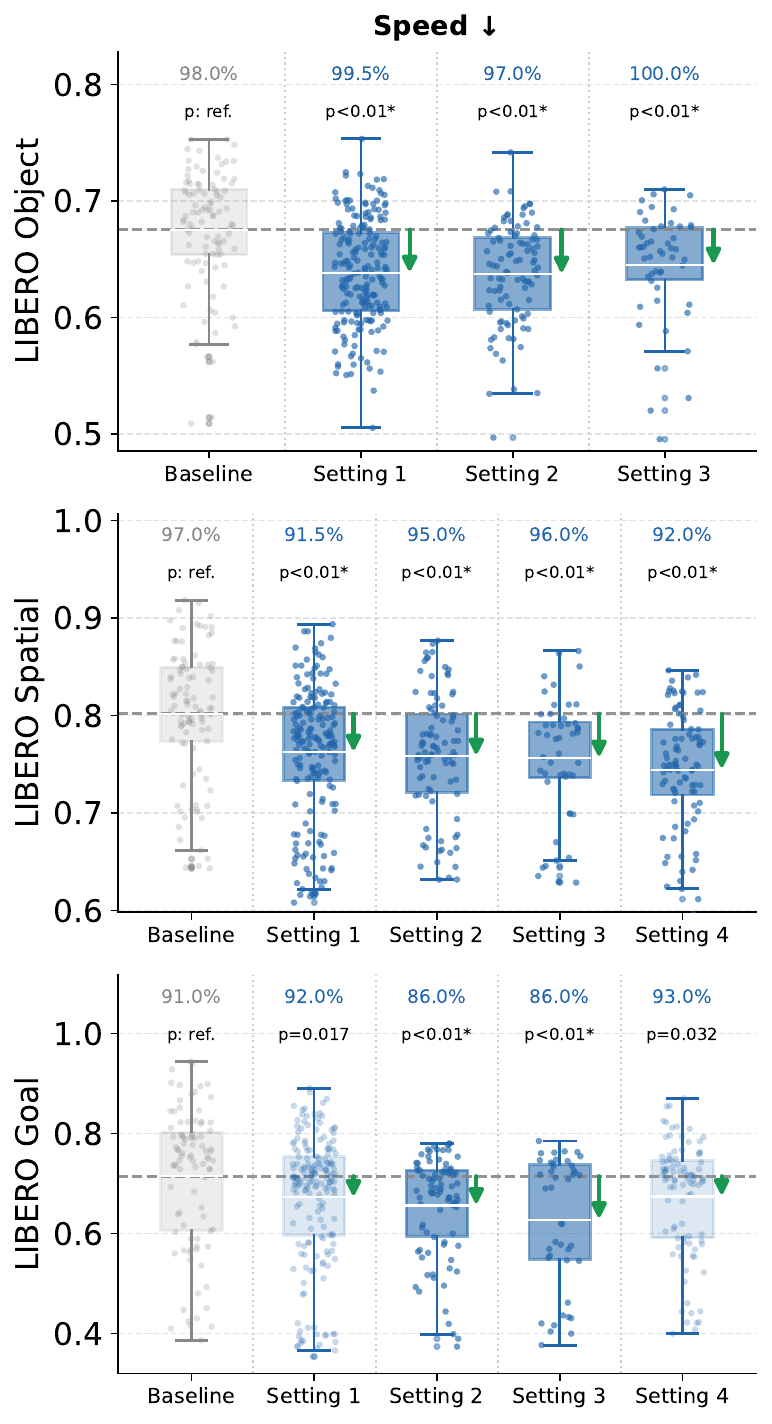}
        \subcaption{$\pi_{0.5}$}
    \end{subfigure}
    \begin{subfigure}{0.48\linewidth}
        \centering
        \includegraphics[width=\linewidth]{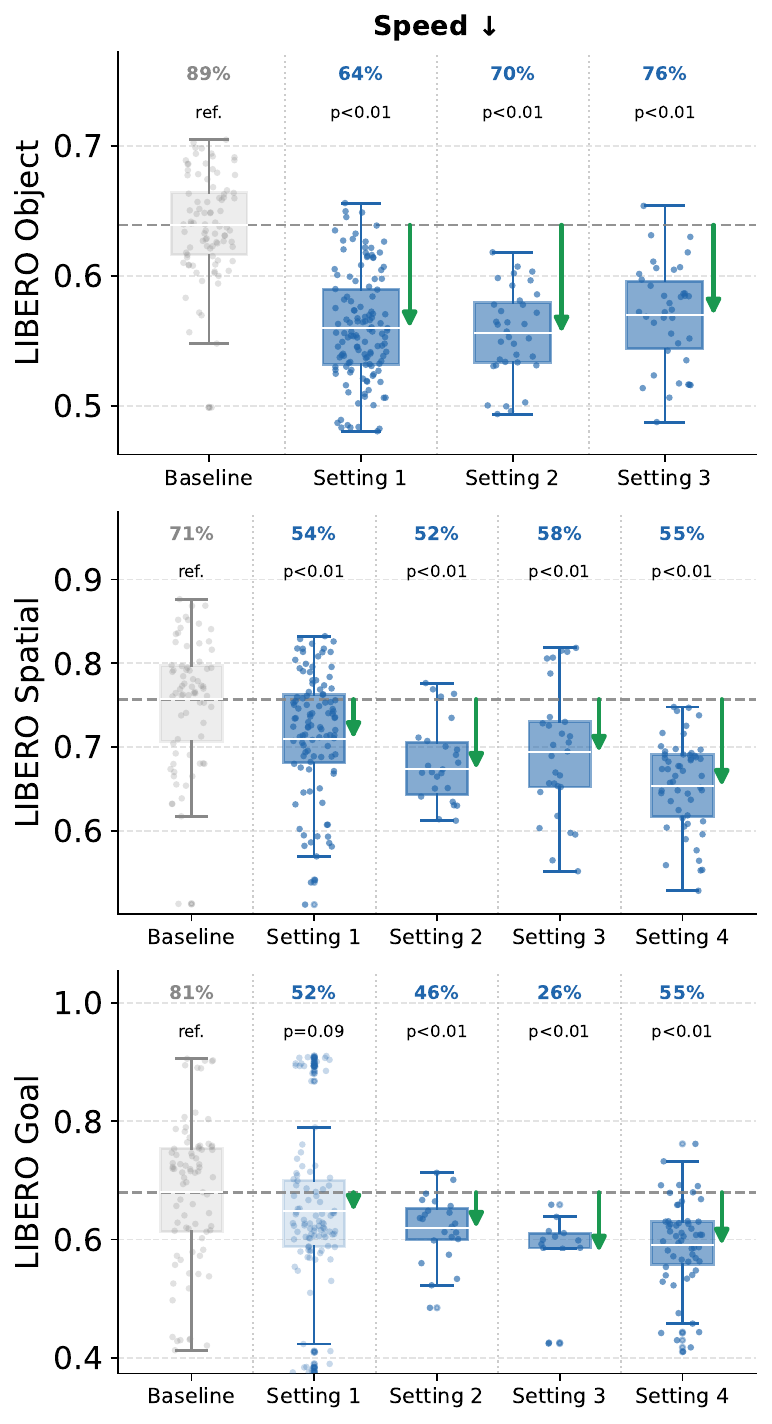}
    
        \subcaption{SmolVLA }
    \end{subfigure}%

    \includegraphics[width=0.5\linewidth]{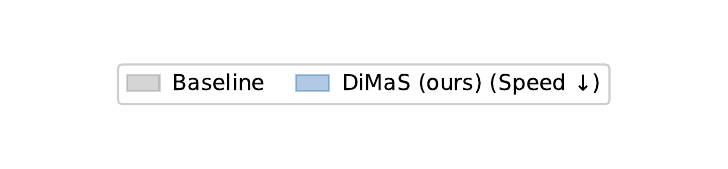}
    \vspace{-2em}
    \caption{\textbf{Steering across evaluation levels on $\pi_{0.5}$ (left) and SmolVLA (right).} We compare the unsteered baseline against DiMaS for decreasing speed. Panels show, left to right, the baseline and the settings discussed in \Cref{gen_des}; columns are the three LIBERO suites (Object, Spatial, Goal). Box plots show the per-episode feature distribution at each level, annotated with success rate and $p$-value vs.\ baseline; box opacity is higher when the shift is statistically significant ($p<0.01$). Arrow length is proportional to the induced $\Delta$mean.}
    \label{fig:levels_comparison_h_to_l_speed}
\end{figure}

\subsection{Steering on longer-horizon tasks.}
\label{sec:extra_settings}
\begin{wrapfigure}{r}{0.42\textwidth}
    \centering
    \vspace{-3\baselineskip}
    \includegraphics[width=0.4\textwidth]{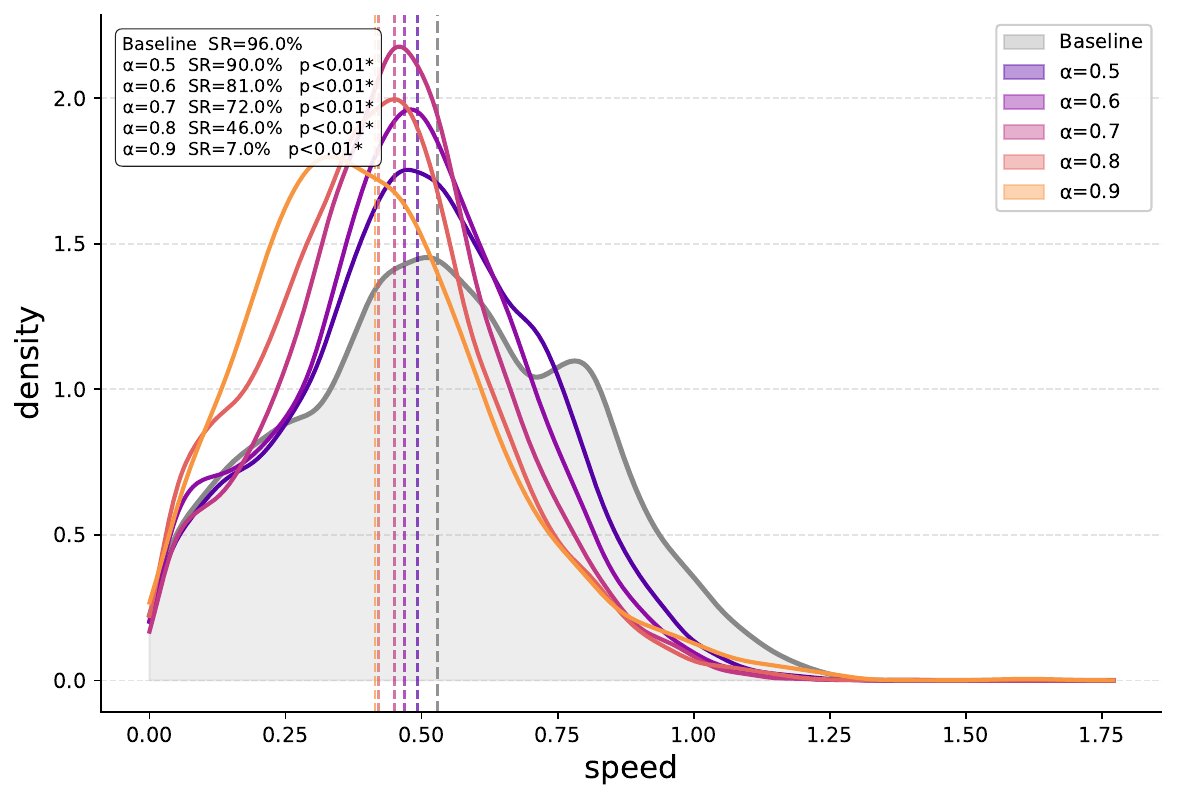}
    \caption{\textbf{Decreasing speed on LIBERO-10 ($\pi_{0.5}$).}}
    \label{fig:libero10_speed}
    \vspace{-3\baselineskip}
\end{wrapfigure}
As shown in \Cref{fig:libero10_speed}, we extend our evaluation to LIBERO-10, whose tasks have substantially longer horizons than the suites above, decreasing speed. DiMaS produces a downward shift ($\Delta\text{mean}=-0.024$, $p<0.01$ for $\alpha=0.5$) at a small success cost ($96\% \to 90\%$), showing that behavioral control remains effective even over extended horizons.
\vspace{-2em}

\section{Investigating internal structure in VLAs}
\label{analysis:representation_structure}

We now analyze representation structure inside the VLM $f_V$ and the action expert $f_A$ to motivate some of DiMaS's design choices and explain why linear steering methods in our experiments are very inconsistent and ineffective. We do so through the lens of output action/trajectory features we would like to control, asking whether each can be represented as a fixed direction in latent space. 
As a running example throughout this section, we investigate SmolVLA \cite{shukor2025smolvla} on LIBERO-Object for the "speed" feature.

\noindent\textbf{Linear separability.} To determine whether a vector/direction in the hidden representation space is a ``good'' candidate for representing a given VLA concept, we first quantify the separability between representations in $\bmD^-, \bmD^+$ across different layers $l$ in the VLM $f_V$, and across different layers $l$ and flow-matching steps $m$ in $f_A$. For $f_V$, we extract the last-token representation $h^{-1}_l(x)$, which summarizes the full input context. To quantify the linear separability, we fit a linear classifier (SVM) on representations $h$ drawn from $\bmD^-$ and $\bmD^+$.

\begin{figure}[H]
        \hspace*{\fill}
        \includegraphics[width=0.24\linewidth]{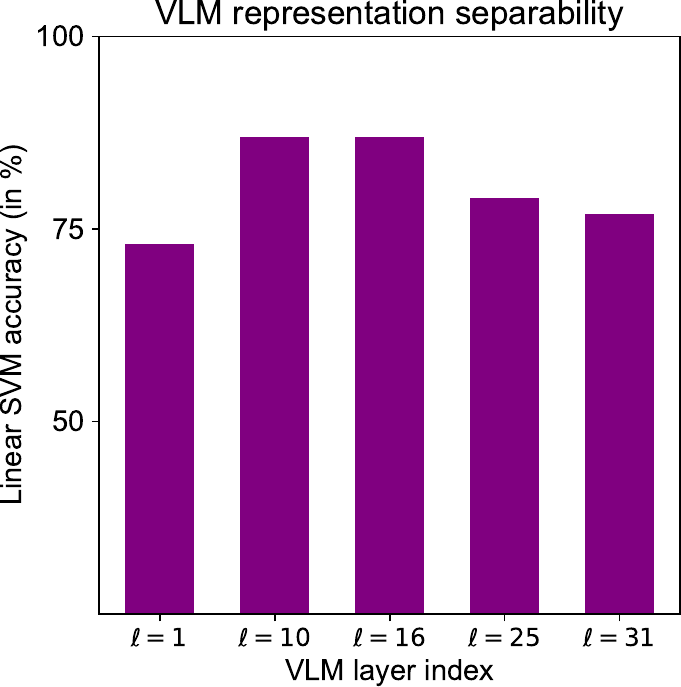}
        \hspace{2pt}
        \includegraphics[width=0.3\linewidth]{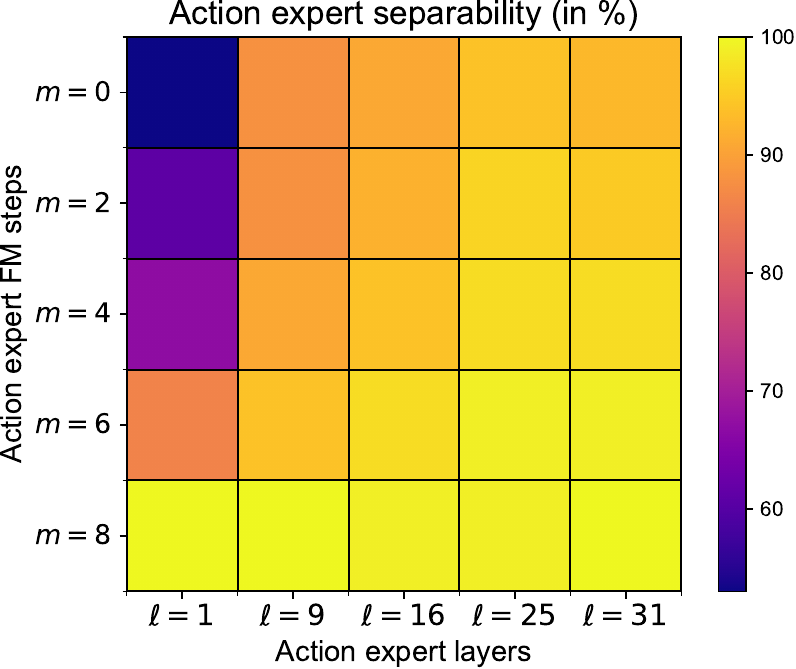}
        \hspace{2pt}
        \includegraphics[width=0.32\linewidth]{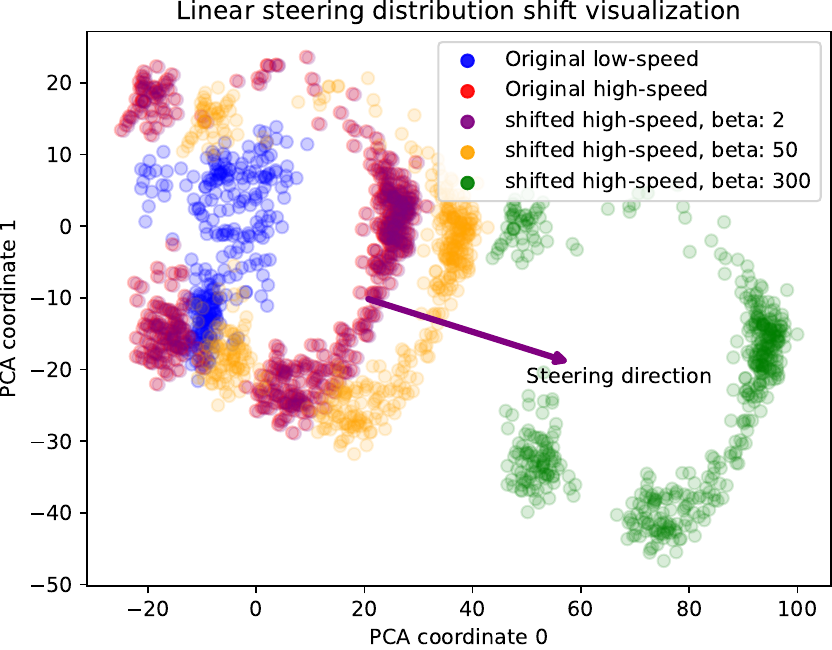}
        \hspace*{\fill}
    \caption{(Left) Linear separability inside SmolVLA for speed feature in different VLM layers and (Middle) different action expert layers, flow-matching steps. Reported as linear SVM classification accuracy ($\%$) between $\bmD^-, \bmD^+$. Higher is better. (Right) 2D PCA visualization for representations in $l=1, m=8$ in action expert. Original high (low) speed representations are in red (blue). Shifting high speed representations along steering vector leads to shifted distributions in orange and green. Linear steering fails to match red and blue distributions.}
    \label{fig:linear_displace_viz}
\end{figure}

We report the separability accuracies in \Cref{fig:linear_displace_viz}. The representations in the action expert are almost completely linearly separable (accuracy near 100\%) at any layer in the later flow-matching steps, as the action tokens are progressively denoised. Interestingly, even at flow-matching step 0, the deeper layers show higher linear separability ($>93\%$) than any layer in the VLM (max 87\%). This is one of the main reasons we choose to intervene in the action expert rather than the VLM, and why we prioritize the deeper layers in the action expert.

\noindent\textbf{Is a linear shift enough to match distributions?}
Even though a linear classifier in the action expert can separate high- and low-speed representations, this alone is not sufficient to validate whether the normal vector to the hyperplane (i.e., the steering vector) represents the feature/concept. To do so, we visualize whether shifting representations along the steering vector transports them from $\bmD^-$ to $\bmD^+$.

We qualitatively visualize this shift in the 2D PCA projection space for layer $l=0$ and flow-matching step $m=8$ in \Cref{fig:linear_displace_viz} (separability is 100$\%$). We shift the high-speed representations (red points) by the unit-norm steering vector scaled by a factor $\beta \in \{2, 50, 300\}$ that controls the strength of linear steering. Note that the representations are steered in the original high-dimensional space. If $\bmD^-, \bmD^+$ can be matched in the original high-dimensional space through a shift in a fixed steering direction, the match will be preserved in the 2D PCA projection space (since PCA projection is a linear operation). \newline
As one might expect, linear steering only shifts the source distribution (red points) in a fixed direction, which leads to the shifted (orange/green) distributions. Since the "shape" of original red (high-speed) and blue (low-speed) point distributions differs, this operation is unable to match the two distributions. It is also worth noting that if $\beta$ is chosen as a small value or as the minimum positive shift needed to flip the classifier's decision (similar to our regression baseline \cite{buurmeijer2026observing}), the shifted distributions nearly coincide with the original red points (purple points mixed with red). These observations indicate that using a steering vector as the concept representation fails to capture the difference between the distributions of high and low speed representations and that we need more complex transformations to move from one to the other.

\vspace{-3em}

\section{Analyzing DiMaS}
\label{sec:analyzing_dimas}
We present two further analyses of DiMaS: the effect of the interpolation factor $\alpha$ on the trade-off between steering strength and task success, and qualitative examples of steered trajectories for reducing vertical displacement. \newline
\label{sec:alpha_ablation}
\begin{center} \noindent 
\begin{minipage}[t]{0.56\textwidth} 
\vspace{0pt} 
\noindent\textbf{Interpolation factor $\alpha$} \newline As described in Section~\ref{method}, our method applies the optimal-transport displacement scaled by an interpolation coefficient $\alpha$, which sets the strength of the intervention. For both $\pi_{0.5}$ and SmolVLA, we vary $\alpha$ and measure the resulting trade-off between success rate (SR) and the magnitude of the induced speed change. 
Figure~\ref{fig:tradeoff_combined} shows that both models follow a broadly similar trend: larger values of $\alpha$ reduce mean speed at the cost of success rate. $\pi_{0.5}$ is the more robust of the two, with success rate dropping only beyond $\alpha = 0.5$, whereas SmolVLA degrades earlier and less regularly. Intermediate values such as $\alpha = 0.5$ offer a favourable trade-off, substantially reducing speed while largely preserving task success, which motivates our choice of $\alpha = 0.5$ as the default operating point throughout the experiments. We provide speed density plots for this ablation on the other LIBERO suites in Appendix~\ref{app:ablation_alpha}, and also ablate the steering layer in Appendix~\ref{app:ablation_layer}. 
\end{minipage} 
\hfill 
\begin{minipage}[t]{0.40\textwidth} \vspace{0pt} \centering \captionsetup{type=figure} \includegraphics[width=0.9\linewidth]{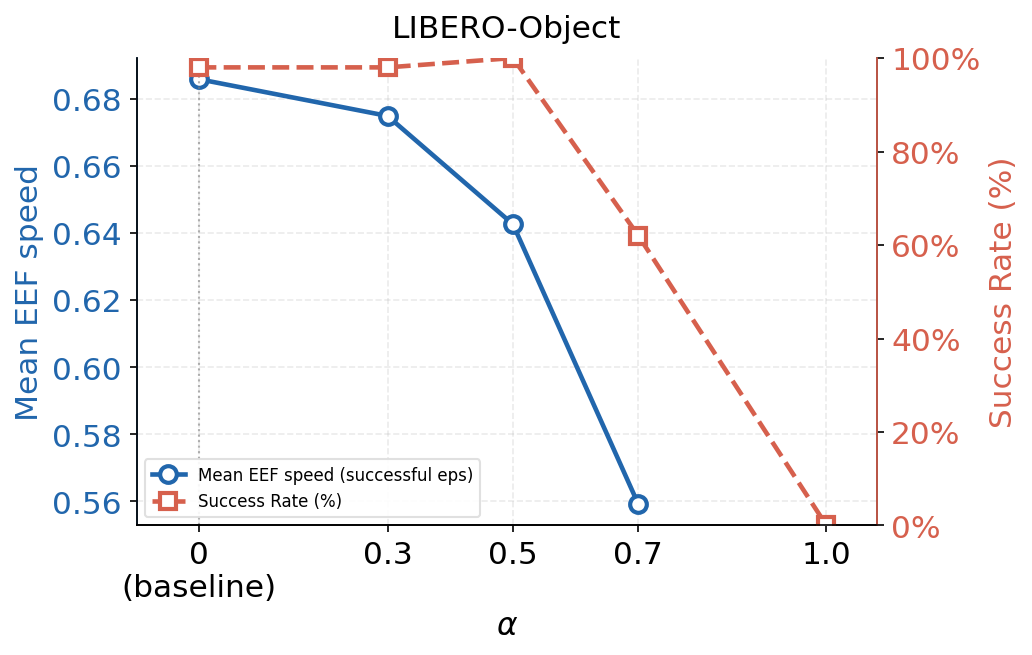} \\[0.3em] {\footnotesize (a) $\pi_{0.5}$} \\[0.6em] \includegraphics[width=0.9\linewidth]{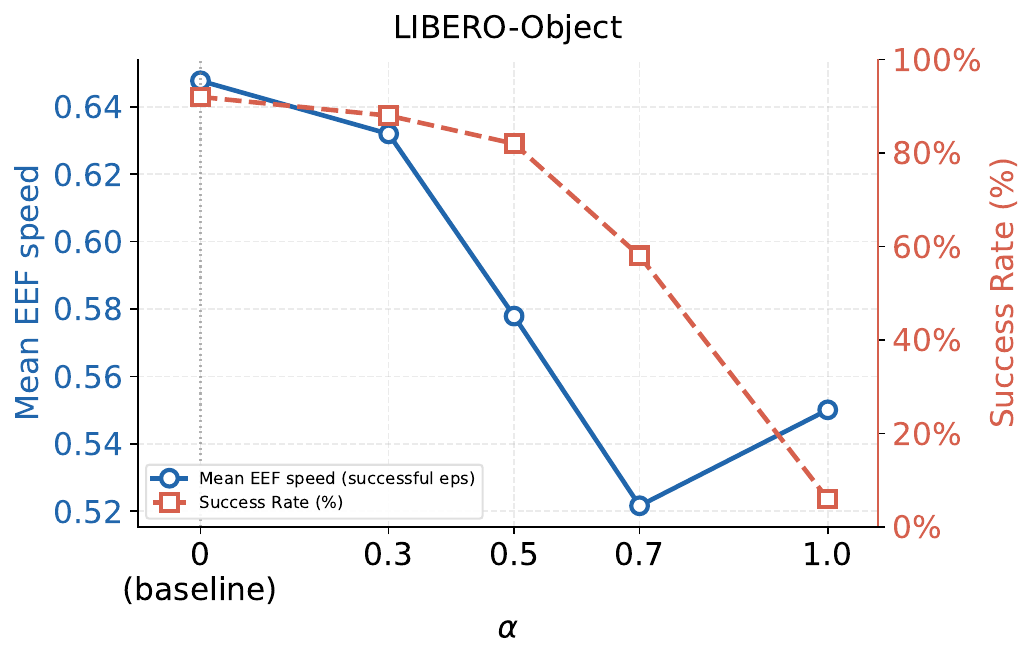} \\[0.3em] {\footnotesize (b) SmolVLA} \captionof{figure}{Trade-off between success rate (SR) and mean end-effector speed as a function of $\alpha$. 
} \label{fig:tradeoff_combined} \end{minipage} \end{center}

\label{sec:qualitative}

\begin{figure}[t]
\centering
\noindent
\begin{minipage}[c]{0.50\textwidth}
    \centering
    \includegraphics[width=0.8\linewidth]{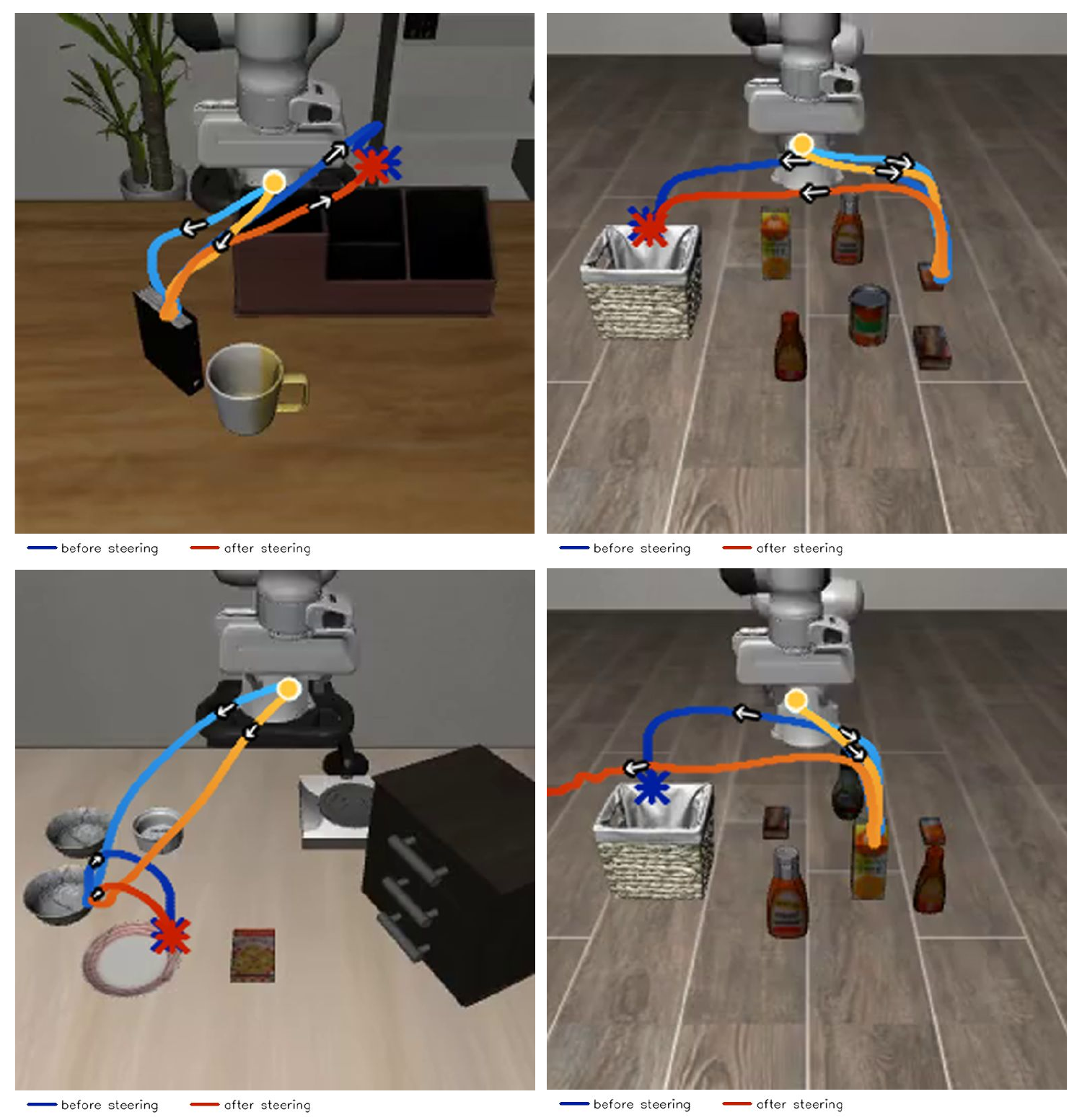}
    \captionof{figure}{Qualitative results for decreasing vertical displacement.}
    \label{fig:lowering_height_examples}
\end{minipage}
\hfill
\begin{minipage}[c]{0.46\textwidth}
    \vspace{-2em}
    \noindent\textbf{Qualitative results} \newline
    Figure~\ref{fig:lowering_height_examples} shows example end-effector trajectories under
    DiMaS steering to decrease vertical displacement, across several LIBERO suites. Comparing
    the original (blue) and steered (red) trajectories, the steered runs consistently exhibit
    smaller vertical displacement, confirming that the intervention produces the intended
    change in executed behavior. The bottom-right example is a failure case, where suppressing
    vertical motion prevents the task from completing successfully.
\end{minipage}
\end{figure}

\section{Conclusion and future work} 
We study behavioral control in Vision-Language-Action (VLA) models: not only \emph{what} a robot does but \emph{how} it does it. Across two state-of-the-art VLAs with a flow-matching action expert, we show that classical linear steering is too restrictive to control behavior reliably, and our analysis of the hidden representation space, in both the vision-language backbone and the action expert, explains why. We address this with \textbf{DiMaS}, which steers behavior by transporting hidden representations from a source distribution, where the target feature is absent, to a target distribution where it is present, using optimal
transport, regularized by an interpolation knob. DiMaS substantially outperforms linear and prompt-based baselines in controlling speed and height while better preserving task success. 
Finally, we evaluate its generalizability across four settings that vary both the diversity of the tasks the steering is learned from and how far the evaluation tasks depart from them, 
and distill from this what an effective steering needs in order to be constructed and applied.
\newline
Our results open several natural extensions. \textit{First}, our intervention currently treats all timesteps alike; a next step is to make \emph{when} to intervene as much a target of learning as \emph{how}, steering selectively (\emph{e.g.}\ only while the end-effector is translating rather than purely rotating) for even smoother control. \textit{Second}, for policies that emit a chunk of actions, such as $\pi_{0.5}$, we associate each hidden representation with its corresponding output action; yet an output action is shaped by more than one representation in the chunk, so grouping representations by the feature of their exact corresponding action is only an approximation, and a finer account of this correspondence could sharpen the learned transport. \textit{Third}, because DiMaS applies to any feature observable from the representations, the same mechanism extends readily to new tasks, embodiments, and more abstract behavioral features, pointing toward representation-level control as a general interface for shaping robot behavior.

\section*{Acknowledgements}
This work has been partially supported by ANR grant VISA DEEP (ANR-20-CHIA-0022), HPC resources of IDRIS under the file A0191016602 allocated by GENCI, and Cluster PostGenAI@Paris (ANR-23-IACL-0007, FRANCE 2030). We also thank Mustafa Shukor for his insightful discussions and valuable feedback.

\bibliographystyle{plainnat}
\bibliography{references}  %

\title{Supplementary Material for \textit{DiMaS}: \textit{Di}stribution \textit{Ma}tching for \textit{S}teering Vision-Language-Action Models}
\maketitle
\raggedbottom
\setcounter{figure}{0}
\renewcommand{\thefigure}{A\arabic{figure}}
\setcounter{table}{0}
\renewcommand{\thetable}{A\arabic{table}}
\setcounter{equation}{0}
\renewcommand{\theequation}{A\arabic{equation}}

We provide details of our proposed method, \textit{DiMaS}, in \Cref{app:dimas}, and describe the baseline methods in \Cref{app:baselines}. \Cref{app:generalization} presents additional experimental results omitted from the main paper. In \Cref{app:analysis_dimas} we analyze DiMaS further, studying the contribution of the interpolation factor $\alpha$ (\Cref{app:ablation_alpha}) and showing how increasing and decreasing vertical displacement affect the end-effector height (\Cref{app:qualitative}). Finally, we ablate the choice of steering layer in \Cref{app:ablation_layer}.

\appendix

\section{Implementation Details of DiMaS}
\label{app:dimas}

\subsection{Steering as a transport map: details}
\label{app:dimas_ot_details}

We provide additional details on the optimal transport formulation introduced in Section 3.2. 
The optimization ranges over $\Pi(\mathcal{D}^{-}, \mathcal{D}^{+})$, the set of all joint distributions (transport plans) on $\mathcal{Z} \times \mathcal{Z}$ whose marginals coincide with $\mathcal{D}^{-}$ and $\mathcal{D}^{+}$. Intuitively, $\gamma(z^{-}, z^{+})$ specifies the allocation of mass from $\mathcal{D}^{-}$ to $\mathcal{D}^{+}$ that minimizes the total quadratic transport cost.

In practice, $\mathcal{D}^{-}$ and $\mathcal{D}^{+}$ are only accessible through finite empirical samples $\mathbf{X}^{-} = \{z^{-}_i\}_{i=1}^{n} \sim \mathcal{D}^{-}$ and $\mathbf{X}^{+} = \{z^{+}_j\}_{j=1}^{m} \sim \mathcal{D}^{+}$. The continuous objective thus reduces to the discrete optimal transport problem,
\begin{equation}
    \min_{\mathcal{T} \in \,\Pi(\mathbf{a},\, \mathbf{b})}
    \left\langle \mathbf{C},\, \mathcal{T} \right\rangle_{\!F} ,
    \label{eq:ot_discrete}
\end{equation}
where $\mathbf{C}_{ij} = \|z^{-}_i - z^{+}_j\|^2$ is the squared Euclidean cost matrix, $\mathbf{a} = \tfrac{1}{n}\mathbf{1}_n$ and $\mathbf{b} = \tfrac{1}{m}\mathbf{1}_m$ are uniform marginal weights, and
\begin{equation*}
    \Pi(\mathbf{a}, \mathbf{b})
    \;=\;
    \Bigl\{
        \mathcal{T} \in \mathbb{R}^{n \times m}_{+}
        \;\Big|\;
        \mathcal{T}\,\mathbf{1}_m = \mathbf{a},\;
        \mathcal{T}^{\!\top}\mathbf{1}_n = \mathbf{b}
    \Bigr\}
\end{equation*}
is the transport polytope. To scale to the high-dimensional hidden representations encountered in VLA models, we solve Eq.~\eqref{eq:ot_discrete} using the \emph{low-rank Sinkhorn} algorithm \cite{scetbon2021lowrank}, which factorises the transport plan as 
\begin{equation}
    \mathcal{T}
    \;=\;
    \mathbf{Q}\;
    \operatorname{diag}(\mathbf{g}^{-1})\;
    \mathbf{R}^{\!\top}
    \label{eq:lowrank}
\end{equation}

\noindent where $\mathbf{Q} \in \mathbb{R}^{n \times r}_{+}$, $\mathbf{R} \in \mathbb{R}^{m \times r}_{+}$, and $\mathbf{g} \in \mathbb{R}^{r}_{+}$ are low-rank factors of rank $r \ll \min(n,m)$.

\noindent Moreover, a regularization term is added to the discrete OT formulation 
\begin{equation}
    \min_{\mathcal{T} \in \,\Pi(\mathbf{a},\, \mathbf{b})}
    \left\langle \mathbf{C},\, \mathcal{T} \right\rangle_{\!F}
    \;-\;
    \varepsilon \,H(\mathcal{T}),
    \label{eq:ot_regularised}
\end{equation}

\noindent where $H(\mathcal{T}) = -\sum_{ij} \mathcal{T}_{ij} \log \mathcal{T}_{ij}$ is the 
entropy of the transport plan and $\varepsilon > 0$ controls the 
degree of regularisation. Smaller $\varepsilon$ gives a sparser and
more faithful approximation of the exact OT solution. We used the implementation of this algorithm in the Python Optimal Transport (POT) package \cite{flamary2021pot}.

\subsection{Hyperparameters and design choices}

The complete set of hyperparameters and design choices involved in DiMaS are given below in \Cref{tab:hyperparams}. The overall design is robust in the sense that all these choices are frozen for all our experiments, which includes all the various model, suite and steering task choices. 

Note that the steering intervention is applied only at the output of a single layer in the action expert, but for all flow-matching steps. $\alpha$ is the most influential choice in terms of balancing control strength with success rate. While $\alpha=0.5$ tends to be a safe default choice, it could be tuned for each task individually for optimal balance. 
The only hyperparameter that depends on the underlying VLA model, and may not generalize to an arbitrary new base VLA, is the layer choice $\ell$.Empirically, we find that late layers work well in general: we use the second-to-last layer, but observe little variation in steering effectiveness across the last few layers. We provide an analysis of this choice for $\pi_{0.5}$ in \Cref{app:ablation_layer}. A more systematic study of layer selection, and of how the optimal layer depends on the base VLA, remains a valuable direction for future work.

\begin{table}[H]
\centering
\caption{Hyperparameters of \textsc{DiMaS}.}
\label{tab:hyperparams}
\footnotesize
\setlength{\tabcolsep}{6pt}
\renewcommand{\arraystretch}{1.2}
\begin{tabular}{lll}
\toprule
\textbf{Component} & \textbf{Hyperparameter} & \textbf{Value} \\
\midrule
\multirow{2}{*}{Distributions $\bmD^-, \bmD^+$}
    & Lower quantile $q^-$ & $0.25$ \\
    & Upper quantile $q^+$ & $0.75$ \\
\midrule
\multirow{2}{*}{Distribution classifier $g$} 
    & Model & SVM (linear kernel) \\
    & Regularisation $C$ & $0.1$ \\
\midrule
\multirow{4}{*}{Learning transport map $\mathcal{T}_l^{(m)}$}
    & Algorithm & Low-rank Sinkhorn \\
    & Regularisation $\varepsilon$ & $10^{-4}$ \\
    & Max.\ iterations & $5\,000$ \\
    & Rank $r$ & $\min(n, m)$ (default) \\
\midrule
\multirow{3}{*}{Steering intervention}
    & Interpolation coefficient $\alpha$ & $0.5$ \\
    & Layer $\ell$ & Second last \\
    & FM steps $m$ & All steps \\
\bottomrule
\end{tabular}
\end{table}

\subsection{Models and datasets}
\label{app:models_and_datasets}

We evaluate our methods on two state-of-the-art VLA models. 
SmolVLA ~\cite{shukor2025smolvla} is a compact vision-language-action model built on a 256M-parameter SmolVLM backbone with a flow-matching action expert. We use the publicly available checkpoint \texttt{lerobot/smolvla\_libero}, fine-tuned on the LIBERO benchmark.
\textbf{$\pi_{0.5}$}~\cite{intelligence2025pi05} is a larger VLA model built on a PaliGemma-3B vision-language backbone combined with a 300M-parameter flow-matching action expert. We use the checkpoint 
\texttt{lerobot/pi05-libero}, fine-tuned on the LIBERO benchmark.
Both models generate actions via iterative denoising over 10 
flow-matching steps. One main difference between these two checkpoints is that the LIBERO-SmolVLA one is trained to predict a chunk of 50 at each timestep and use only the first
one at inference time, whereas for $\pi_{0.5}$ it is able to predict and execute 10 actions directly, which makes its trajectories smoother and faster to run.

To evaluate our proposed steering strategy we use LIBERO~\cite{libero}, a benchmark for lifelong robot learning comprising five task suites: \textit{LIBERO-Object}, \textit{LIBERO-Spatial}, \textit{LIBERO-Goal}, \textit{LIBERO-10} (also referred to as \textit{LIBERO-Long}), each containing 10 tasks, and \textit{LIBERO-90}, which contains 90 short-horizon tasks, for a total of 130 manipulation tasks. We evaluate on four suites: 
\begin{itemize} 
\item \textbf{LIBERO-Object} focuses on object-centric manipulation, requiring the robot to pick up and place various household objects. 
\item \textbf{LIBERO-Spatial} introduces spatial reasoning, requiring the robot to reason about object relationships. \item \textbf{LIBERO-Goal} contains goal-conditioned tasks with more diverse objectives, including articulated object manipulation. 
\item \textbf{LIBERO-10} (\textit{LIBERO-Long}) consists of long-horizon tasks that require composing multiple sub-goals in sequence. 
\end{itemize}

\subsection{Computational Cost} \label{app:compute}

Computing the OT transport map for a single suite (LIBERO-Object) 
across 50 training episodes represents approximately $\sim$3{,}500 samples per distribution. Solving the Low-Rank 
Sinkhorn problem across all 10 flow-matching steps takes 
approximately 85 minutes. This cost is paid once 
offline and does not affect inference latency. Theoretically, the time complexity of the Low-Rank Sinkhorn algorithm 
is $\mathcal{O}\bigl((n + m) \cdot r \cdot K\bigr)$, where $n$ and 
$m$ are the number of source and target samples, $r$ is the rank of 
the factorisation, and $K$ is the number of iterations.

Regarding inference-time intervention, all experiments were conducted on NVIDIA Tesla V100, A100, and H100 GPUs. Our steering mechanism is lightweight: the overhead (SVM classification + transport) averages only $0.62 \pm 0.05$\,ms per flow-matching step where an intervention occurs (calculated on $n=98$ intervention steps). Even in the worst case, where the classifier triggers steering at all 10 flow-matching steps, this amounts to just $6.2$\,ms of overhead per robot timestep. Since each robot timestep corresponds to an action budget of approximately $50$\,ms, this worst-case overhead consumes only ${\sim}12\%$ of the available budget, leaving substantial headroom for real-time deployment.

\section{Baseline Details}
\label{app:baselines}

The representation-level baselines intervene either in the VLM backbone or in the action expert $f_A$. Following prior work~\cite{buurmeijer2026observing}, the VLM interventions are applied at \emph{every} layer of the backbone, using the token-averaged residual stream $\bar{h}_\ell(x_t) = \mathbb{E}_p\!\left[h^p_\ell(x_t)\right] \in \mathbb{R}^{d_V}$ at each layer $\ell$; the action-expert intervention acts at a single layer $\ell$, using the per-token representations $h^{p,m}_\ell(x_t) \in \mathbb{R}^{d_A}$, where $m$ indexes the denoising step. In all cases the additive shift is added to every token position $p$ (and, in the action expert, every denoising step $m$) at the intervened layer(s). \newline

Following the main paper, we split representations into a source distribution $\mathcal{D}^{-}$ (feature value below $q_\tau$) and a target distribution $\mathcal{D}^{+}$ (feature value above $q_{1-\tau}$), referred to as the source and target \emph{tails} of the feature distribution. We write $h \in \mathbb{R}^d$ for a generic representation, with $d \in \{d_V, d_A\}$, and all quantities below are fit and applied per intervened layer. \newline

All representation-level baselines share the feature filter of the main paper: the shift $\mathcal{T}(h)$ is applied only to representations predicted to lack the targeted feature, gated by the binary classifier $g_\ell^{(m)}(h) \in \{0, 1\}$ that is active ($g_\ell^{(m)}(h) = 1$) when $h$ is assigned to $\mathcal{D}^{-}$. The effective update is therefore \begin{equation} h \;\mapsto\; 
\begin{cases} \mathcal{T}(h) & \text{if } g_\ell^{(m)}(h) = 1, \\[4pt] h & \text{otherwise,} \end{cases} \end{equation} 
so that representations already exhibiting the feature are left unchanged. The baselines differ only in the choice of transport map $\mathcal{T}$. \newline

\noindent\textbf{Mean-difference steering (VLM Mean).}
This baseline applies a constant additive shift equal to the difference of the tail means,
\begin{equation}
    \mathcal{T}(h) = h + (\mu^{+} - \mu^{-}), \qquad
    \mu^{\pm} = \mathbb{E}_{h \sim \mathcal{D}^{\pm}}[h],
\end{equation}
with $\mu^{+}$ and $\mu^{-}$ estimated over the target and source tails. \newline

\noindent\textbf{Regression-based steering (VLM Regression, FM Regression).}
This baseline replaces the constant shift with an adaptive, representation-dependent one, and is applied identically to the VLM and action-expert representations. We fit a ridge regressor on the representations to predict the scalar feature,
\begin{equation}
    \hat{\phi}_{\ell}(h) = w_\ell^\top h + b_\ell, \qquad w_\ell \in \mathbb{R}^{d},
\end{equation}

and set a target value $s^\star$ equal to a tail quantile of the feature distribution,
$s^\star \in \{q_\tau, q_{1-\tau}\}$, depending on the steering direction. The
representation is then displaced by the minimum-norm perturbation that moves the predicted
feature from $\hat{s} = \hat{\phi}_{\ell}(h)$ to $s^\star$,
\begin{equation}
    \mathcal{T}(h) = h + (s^\star - \hat{s})\,\frac{w_\ell}{\lVert w_\ell \rVert_2^{2}}.
\end{equation}

\textsc{VLM Regression} intervenes on the token-averaged VLM representation at every layer,
and \textsc{FM Regression} on the action-expert representation at a single layer. \newline

\noindent\textbf{Prompt steering.}
As a non-representational baseline, we steer the target feature through the language
instruction, appending a directive to the task prompt rather than modifying any hidden
representation. For speed we append ``\texttt{Do this quickly.}'' (low-to-high) or
``\texttt{Do this slowly.}'' (high-to-low), and for end-effector height ``\texttt{Do this
at high height.}'' or ``\texttt{Do this at low height.}''

\section{Additional experimental results}
\label{app:generalization}

\begin{figure}[H]
    \centering

    \begin{subfigure}{0.48\linewidth}
        \centering
        \includegraphics[width=\linewidth]{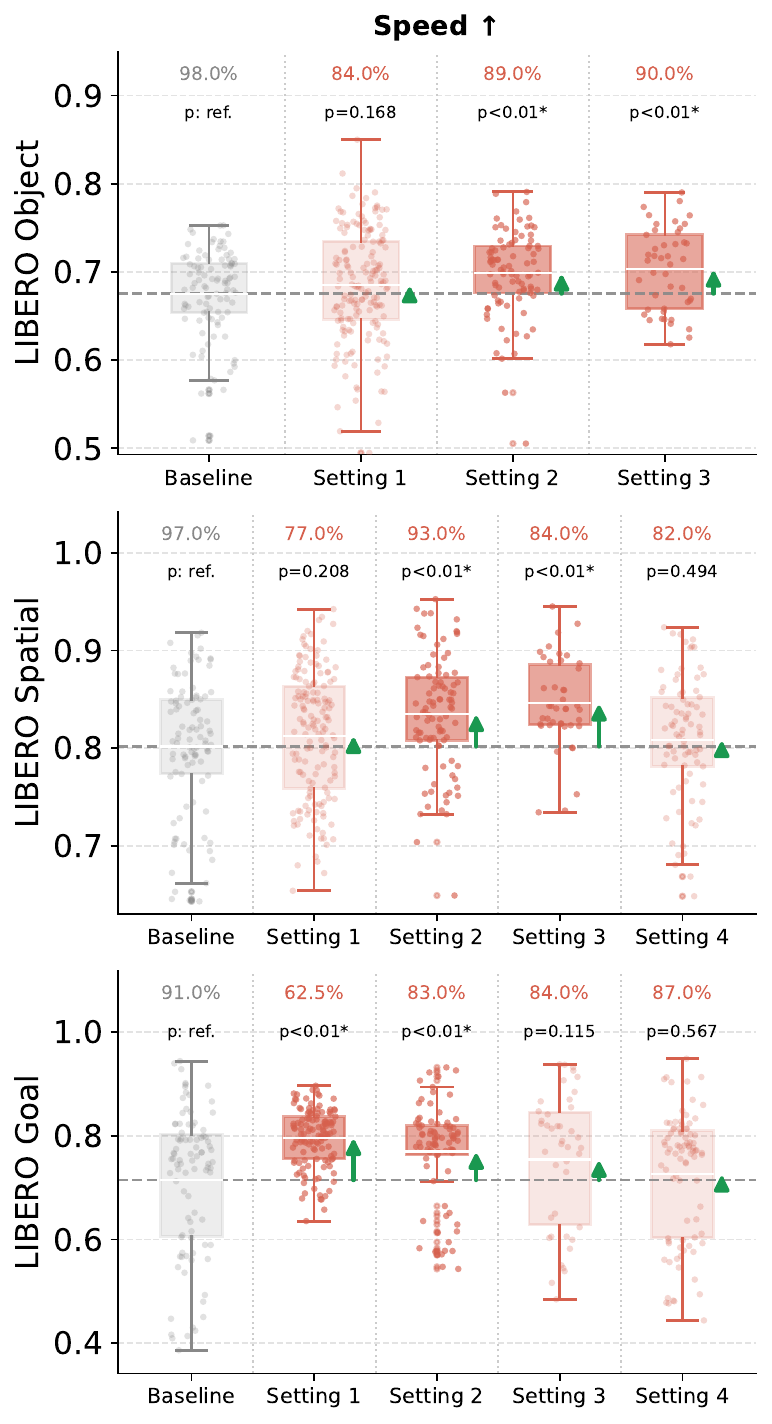}
        \subcaption{$\pi_{0.5}$}
    \end{subfigure}
    \begin{subfigure}{0.48\linewidth}
        \centering
        \includegraphics[width=\linewidth]{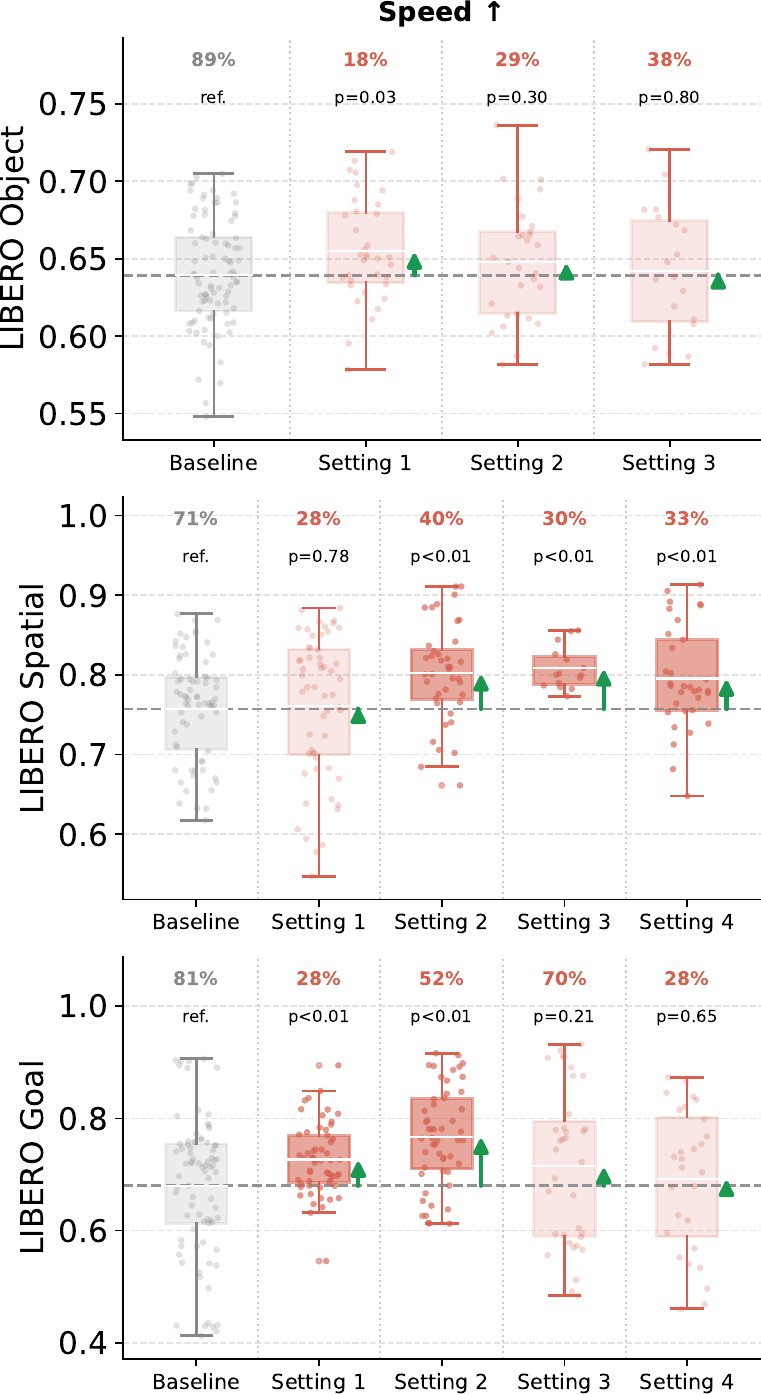}
        \subcaption{SmolVLA }
    \end{subfigure}

    \includegraphics[width=0.5\linewidth]{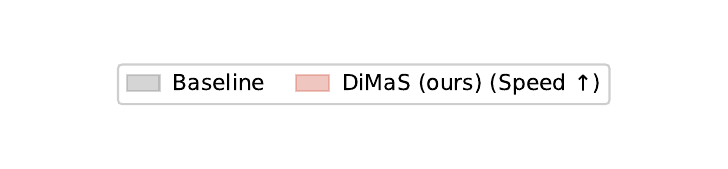}
    \vspace{-2em}

    \caption{\textbf{Steering across evaluation levels on $\pi_{0.5}$ (left) and SmolVLA (right).} We compare the unsteered baseline against DiMaS for increasing speed. Panels show, left to right, the baseline and the settings discussed in Section 4.2; columns are the three LIBERO suites (Object, Spatial, Goal). Box plots show the per-episode feature distribution at each level, annotated with success rate and $p$-value vs.\ baseline; box opacity is higher when the shift is statistically significant ($p<0.01$). Arrow length is proportional to the induced $\Delta$mean.}
    \label{fig:levels_comparison_l2h_speed}
\end{figure}

\begin{figure}[H]
    \centering

    \begin{subfigure}{0.48\linewidth}
        \centering
        \includegraphics[width=\linewidth]{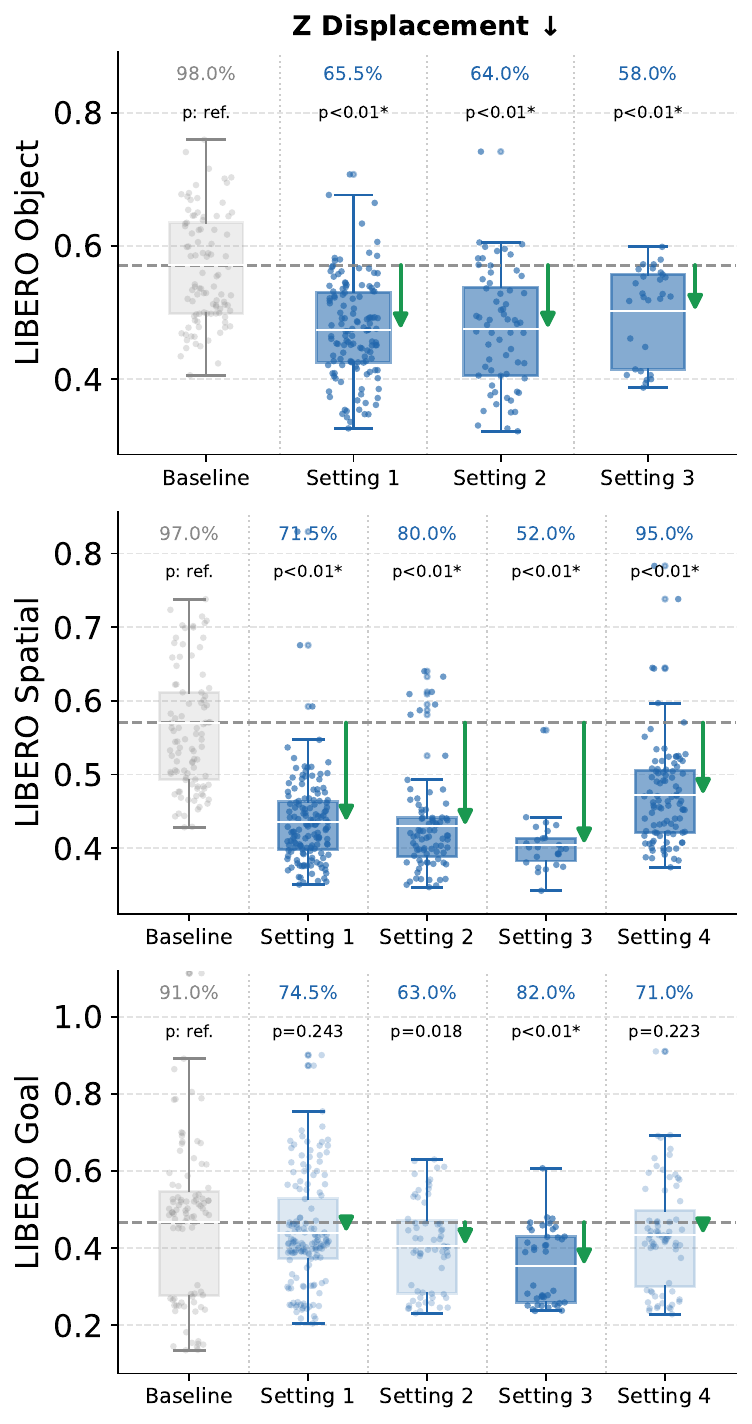}
        \subcaption{$\pi_{0.5}$}
    \end{subfigure}%
    \begin{subfigure}{0.495\linewidth}
        \centering
        \includegraphics[width=\linewidth]{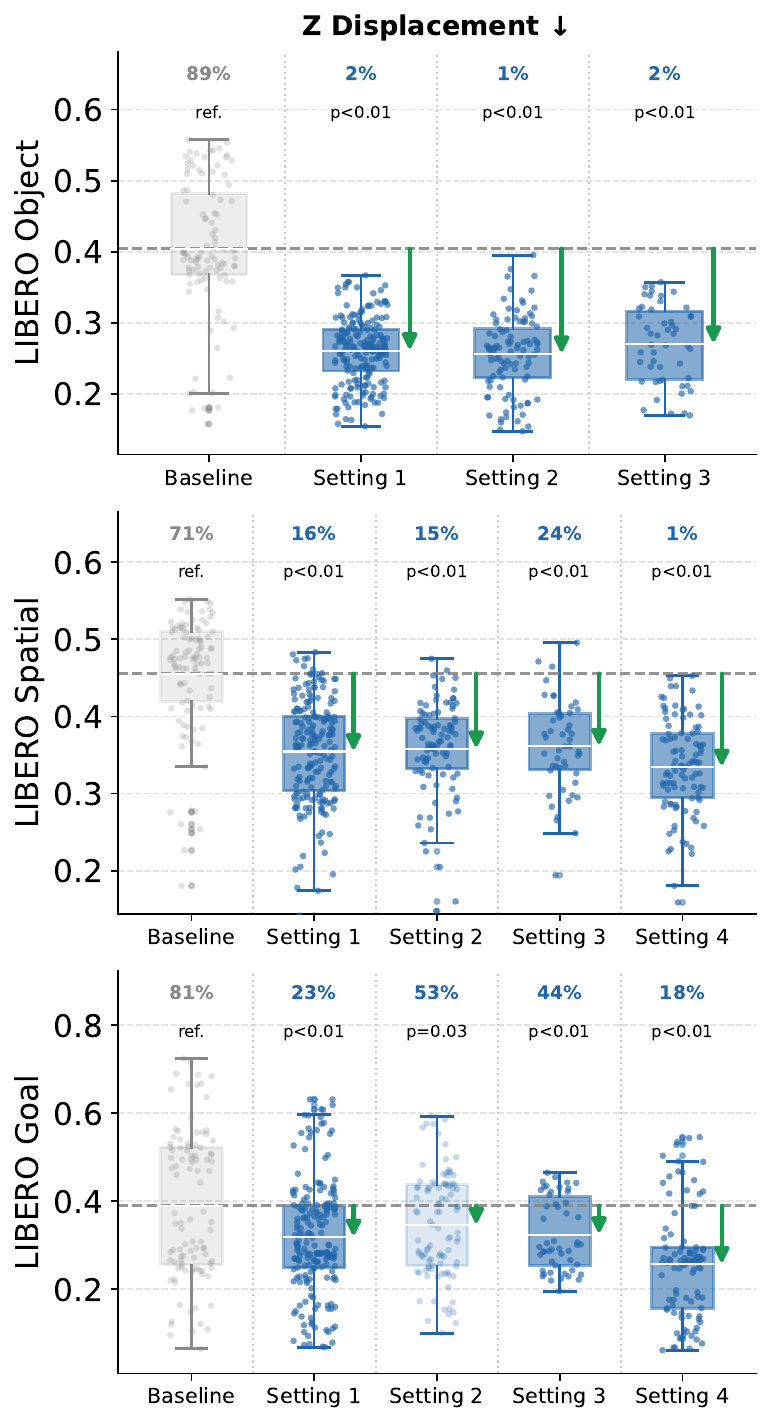}
        \subcaption{SmolVLA }
    \end{subfigure}%

    \includegraphics[width=0.5\linewidth]{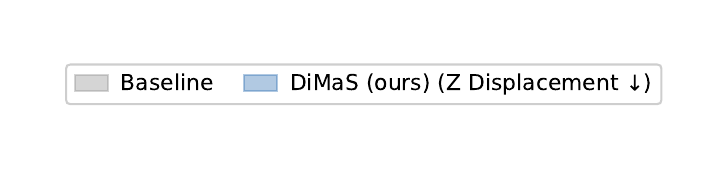}
    \vspace{-2em}

    \caption{\textbf{Steering across evaluation levels on $\pi_{0.5}$ (left) and SmolVLA (right).} We compare the unsteered baseline against DiMaS for decreasing vertical displacement. Panels show, left to right, the baseline and the settings discussed in Section 4.2; columns are the three LIBERO suites (Object, Spatial, Goal). Box plots show the per-episode feature distribution at each level, annotated with success rate and $p$-value vs.\ baseline; box opacity is higher when the shift is statistically significant ($p<0.01$). Arrow length is proportional to the induced $\Delta$mean.}
    \label{fig:levels_comparison_h2l_height}
\end{figure}

\begin{figure}[H]
    \centering

    \begin{subfigure}{0.48\linewidth}
        \centering
        \includegraphics[width=\linewidth]{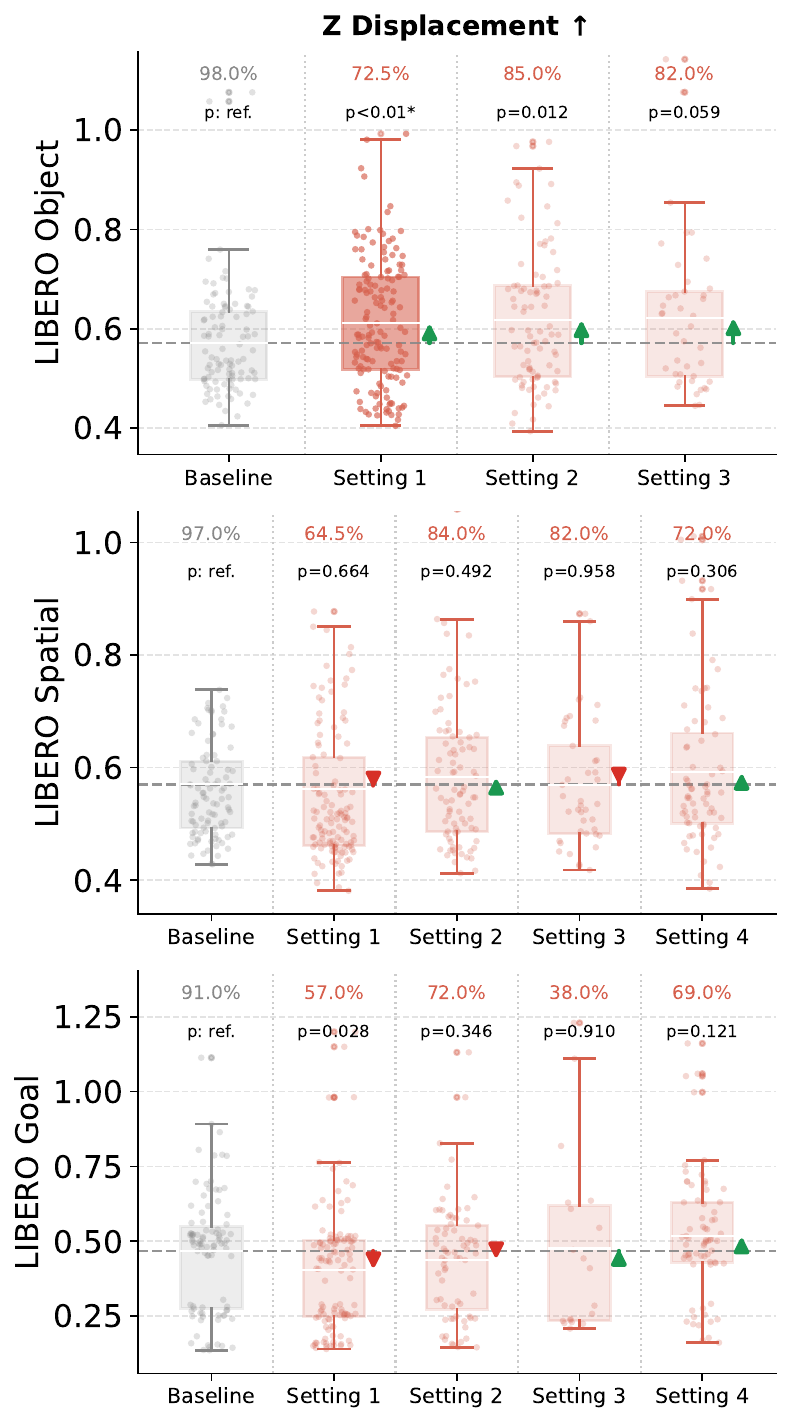}
        \subcaption{$\pi_{0.5}$}
    \end{subfigure}%
    \begin{subfigure}{0.47\linewidth}
        \centering
        \includegraphics[width=\linewidth]{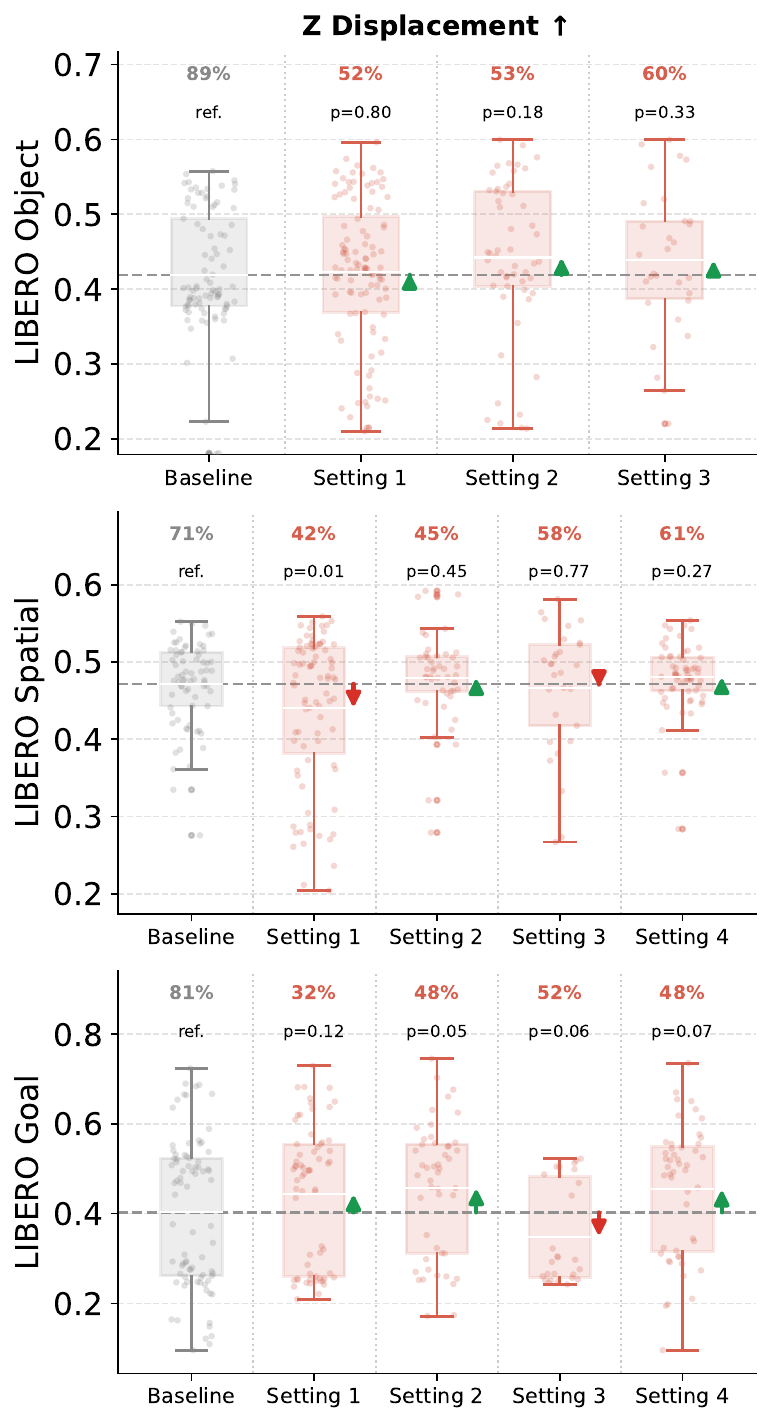}
        \subcaption{SmolVLA }
    \end{subfigure}%

    \includegraphics[width=0.5\linewidth]{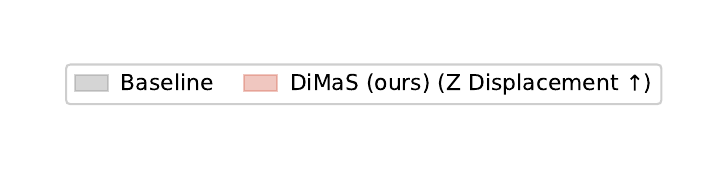}
    \vspace{-2em}

    \caption{\textbf{Steering across evaluation levels on $\pi_{0.5}$ (left) and SmolVLA (right).} We compare the unsteered baseline against DiMaS for increasing vertical displacement. Panels show, left to right, the baseline and the settings discussed in Section 4.2; columns are the three LIBERO suites (Object, Spatial, Goal). Box plots show the per-episode feature distribution at each level, annotated with success rate and $p$-value vs.\ baseline; box opacity is higher when the shift is statistically significant ($p<0.01$). Arrow length is proportional to the induced $\Delta$mean.}
    \label{fig:levels_comparison_l2h_height}
\end{figure}

The main paper reports the generalization study for decreasing end-effector speed. Here we provide the remaining three feature and direction combinations across all four settings and both models: increasing speed (\Cref{fig:levels_comparison_l2h_speed}), decreasing vertical displacement (\Cref{fig:levels_comparison_h2l_height}), and increasing vertical displacement (\Cref{fig:levels_comparison_l2h_height}).

\noindent\textbf{Decreasing vertical displacement.} This direction closely mirrors the speed-decrease result of the main paper (\Cref{fig:levels_comparison_h2l_height}). On both models, the intervention yields a significant downward shift at every setting on Object and Spatial ($p<0.01$), including transfer to held-out tasks (Setting~3) and, for Spatial, to a disjoint suite (Setting~4). As with speed, LIBERO-Goal is the harder case: the shift stays in the intended direction but reaches significance only at some settings ($\pi_{0.5}$ Setting~3, $p<0.01$; SmolVLA Setting~4). The strong effect on Object and Spatial comes with a larger success-rate cost than for speed, consistent with vertical displacement being more tightly coupled to the lifting and placing phases needed to complete the task.

\noindent\textbf{The increasing directions are more constrained by the task.} Steering the two increasing directions is more demanding than the decreasing ones. For increasing speed (\Cref{fig:levels_comparison_l2h_speed}), $\pi_{0.5}$ shifts the feature in the intended direction throughout and is significant at the intermediate settings; the effect weakens only under the largest shift (Setting~4 on Spatial and Goal, $p=0.494$ and $p=0.567$), and on SmolVLA it holds in a subset of cells. Increasing vertical displacement (\Cref{fig:levels_comparison_l2h_height}) is the most constrained combination, with a significant effect concentrated in fewer cells (e.g.\ $\pi_{0.5}$ Object, Setting~1, $p<0.01$). We read this asymmetry as reflecting the policy's default behavior: successful trajectories naturally decelerate and lower the end-effector to grasp and place, so steering that reinforces these tendencies (slower, lower) aligns with the direction the representations already encode, whereas steering the opposite way (faster, higher) pushes against it. Steering is thus most effective when it amplifies a behavior the policy is already inclined toward, rather than reversing it.

\section{Additional DiMaS analysis}
\label{app:analysis_dimas}

We report additional results complementing the main paper: speed density plots for the $\alpha$ ablation across additional task suites for both $\pi_{0.5}$ and SmolVLA in \Cref{app:ablation_alpha}, and examples of end-effector height under increasing and decreasing vertical-displacement steering in \Cref{app:qualitative}.
\subsection{Interpolation factor: per-suite density}
\label{app:ablation_alpha}

\begin{figure}[H]
    \centering

    \begin{subfigure}{0.4\linewidth}
        \centering
        \includegraphics[width=\linewidth]{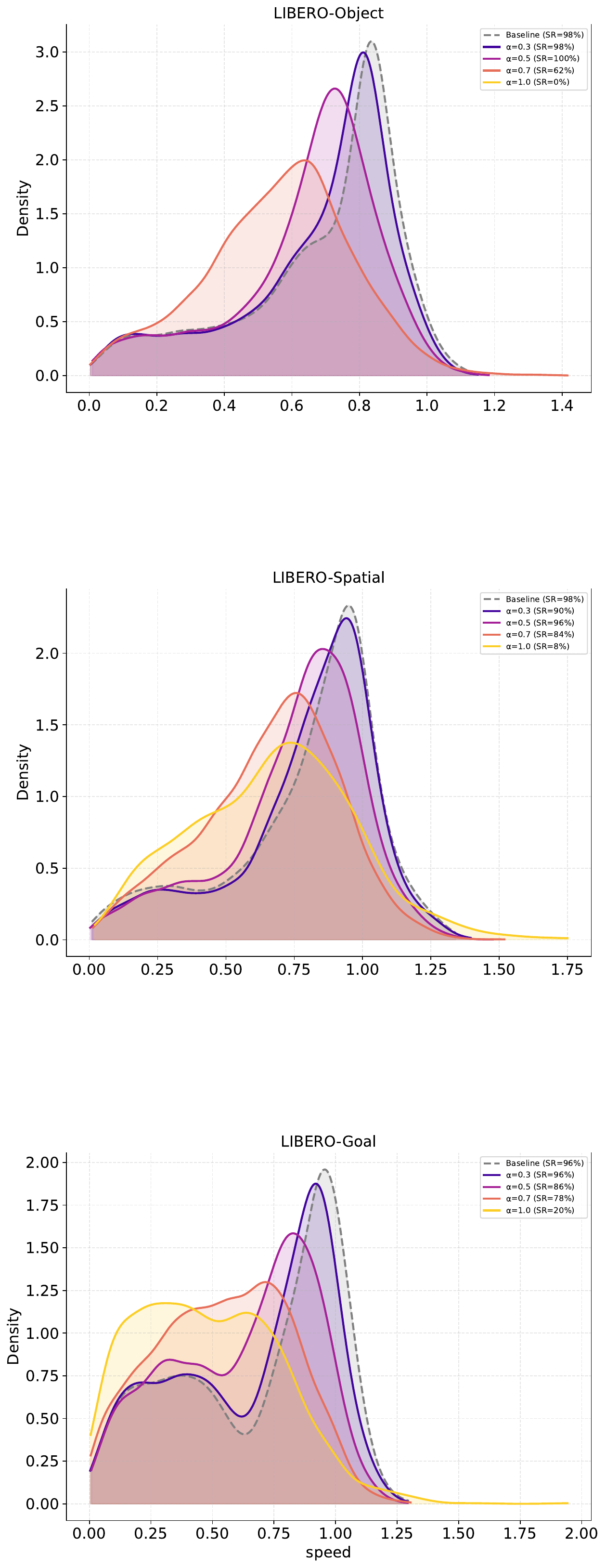}
        \subcaption{$\pi_{0.5}$}
    \end{subfigure}%
    \begin{subfigure}{0.4\linewidth}
        \centering
        \includegraphics[width=\linewidth]{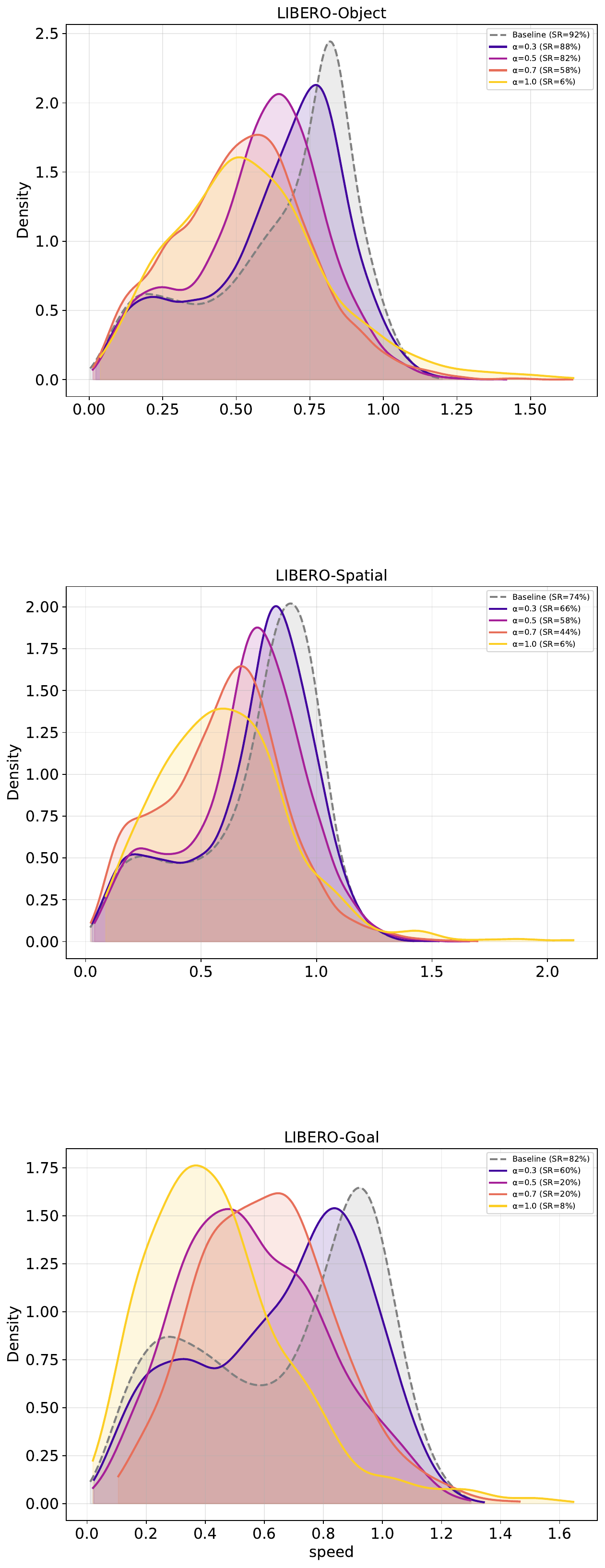}
        \subcaption{SmolVLA }
    \end{subfigure}%

    \caption{\textbf{Effect of the interpolation coefficient $\alpha$ on speed and task success.} Speed distributions when steering towards lower speeds for $\pi_{0.5}$ (top) and SmolVLA (bottom).
    Different curves show the resulting speed distributions when changing the interpolation factor of DiMaS.}
    \label{fig:alpha_ablation_density}
\end{figure}

Figure~\ref{fig:alpha_ablation_density} highlights the clear distribution shift 
introduced by \textsc{DiMaS} for speed H$\to$L steering. Across all three suites, increasing $\alpha$ consistently shifts the distribution towards lower speed values. However, the success rate also drops gradually for $\alpha \in \{0.3, 0.5\}$ and 
more sharply for $\alpha \in \{0.7, 1.0\}$. 

\subsection{Additional qualitative results}
\label{app:qualitative}

\begin{figure}[H]
    \centering
    \includegraphics[width=1.0\linewidth]{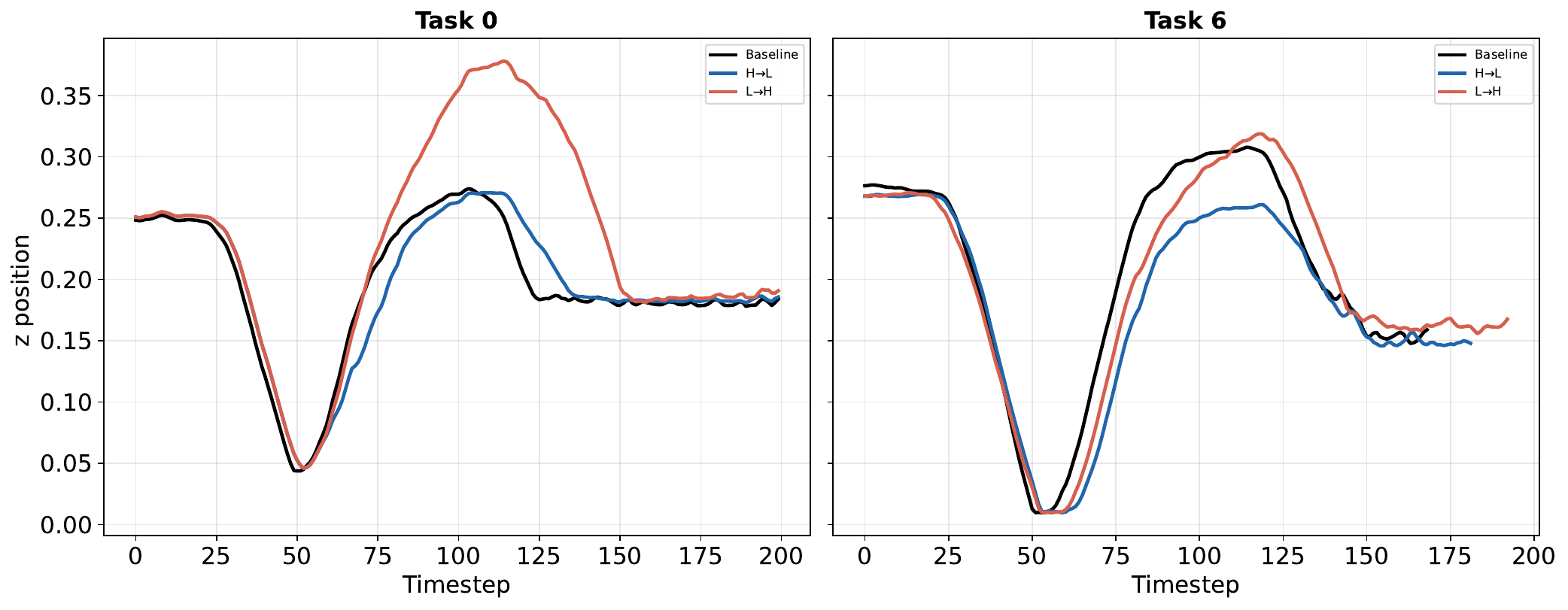}
   \caption{\textbf{End-effector height trajectories under DiMaS height steering on SmolVLA.} Evolution of the observed EEF $z$ position over time for two successful episodes of LIBERO-Object tasks (task 0 and task 6). Curves show the unsteered baseline (black), H$\to$L steering (blue), and L$\to$H steering (red).}
    \label{fig:overtime_steering}
\end{figure}
Figure~\ref{fig:overtime_steering} shows the evolution of the vertical end-effector position over time for two episodes of the LIBERO-Object suite, illustrating the effect of DiMaS height steering on the trajectories. On the left, the red curve (L$\to$H) is consistently higher than the baseline during the transport phase, while the blue curve (H$\to$L) remains close to or below it. On the right, H$\to$L produces a noticeably lower trajectory while L$\to$H follows the baseline closely. This suggests that the steering effect is task-dependent.

\section{Steering-layer ablation}
\label{app:ablation_layer}

\begin{figure}[t]
    \centering
    \begin{subfigure}{\linewidth}
        \centering
        \includegraphics[width=\linewidth]{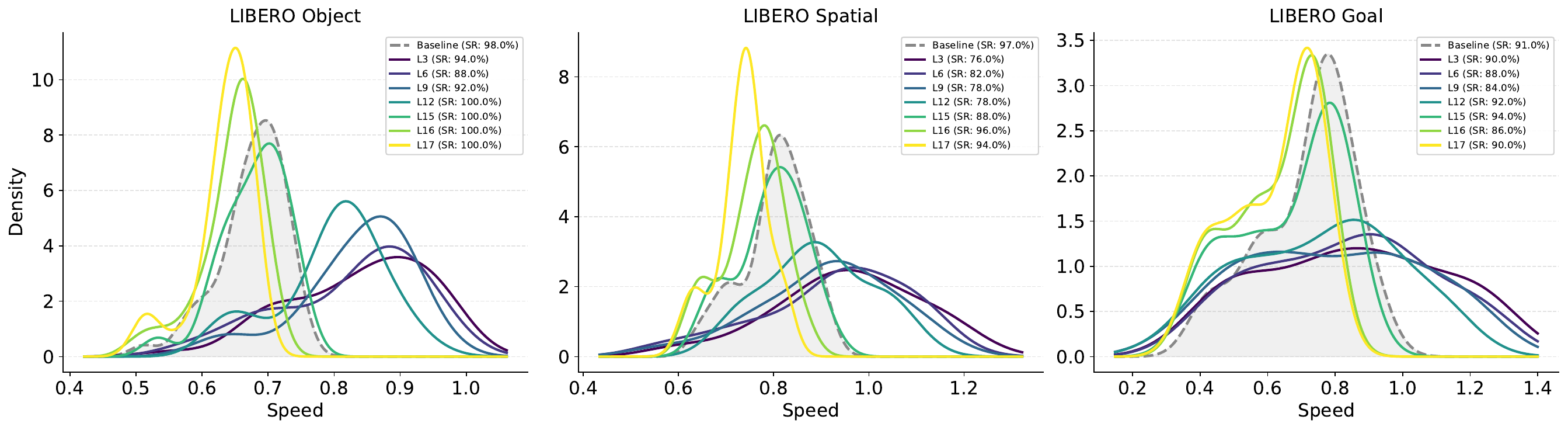}
        \caption{Steering to decrease speed (H$\to$L).}
    \end{subfigure}

    \vspace{0.5em}

    \begin{subfigure}{\linewidth}
        \centering
        \includegraphics[width=\linewidth]{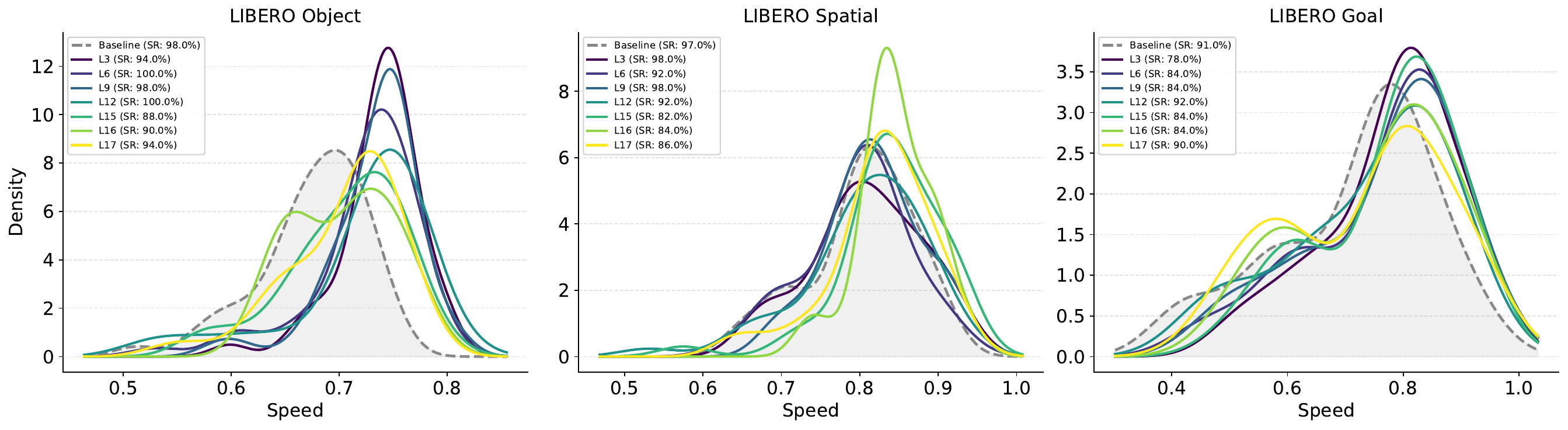}
        \caption{Steering to increase speed (L$\to$H).}
    \end{subfigure}

    \caption{\textbf{Effect of steering layer on speed and task success.} Results for $\pi_{0.5}$ across the three LIBERO suites. Each curve shows the resulting speed distribution when steering at a different layer (L3--L17), with the corresponding success rate (SR) in the legend; the shaded region is the unsteered baseline.}
    \label{fig:layer_ablation}
\end{figure}

Figure~\ref{fig:layer_ablation} shows that the choice of steering layer strongly determines the induced speed shift and, to a lesser extent, the preserved success rate. For decreasing speed (top row), steering at later layers (L15--L17) most effectively concentrates mass at lower speeds, moving the distribution below the baseline while largely preserving task success. Earlier and middle layers (L3--L9) are markedly less effective: they shift mass toward higher speeds, opposite to the intended direction. The accompanying drop in success rate (e.g.\ SR $=76\%$ on Spatial at L3) partly reflects this loss of control, but also the faster execution itself, since moving more quickly leaves less margin to grasp and place the object reliably. The same layer dependence holds for increasing speed (bottom row), where later layers again produce the cleanest shift in the intended direction. Across both directions and all three suites, the behavioral feature is most cleanly and controllably represented in the later layers of the network, which motivates our choice of a late intervention layer in the main experiments.

\end{document}